\documentclass[journal]{IEEEtran}
\ifCLASSINFOpdf
\else
\fi
	\usepackage{xcolor}
	
	\usepackage{cite}
	\usepackage{amsmath,amssymb,amsfonts}
	\usepackage{algorithmic}
	\usepackage{graphicx}
	\usepackage{textcomp}
	\usepackage{url}
	\usepackage[caption=false,font=footnotesize]{subfig}
	\usepackage{tikz}
	\usepackage{pgfplots}
	\usepackage{tabularx}
	\usepackage{makecell}

	\usetikzlibrary{calc}
	\usepgfplotslibrary{groupplots}
	\pgfplotsset{compat=1.18}
	\usepackage[hidelinks]{hyperref}
	\usepackage{orcidlink}

	\newcommand{\rev}[1]{#1}
	\newcommand{\majorrev}[1]{#1}
	
	\newcommand{\compacttablebottom}{\vspace{-0.5em}}

	\usepackage{multicol}
	\usepackage{fancyhdr}
	\fancypagestyle{firstpage}{% Page style for first page
		\fancyhf{}% Clear header/footer
		\fancyhead[C]{\small{This work has been submitted to the IEEE for possible publication. \\
				Copyright may be transferred without notice, after which this version may no longer be accessible.}}% Header
	}

\begin{document}
		
		% \title{
			%   A Reinforcement Learning Framework \\ for Near-Zero Violation Shared Control in Surgical Tasks
			% }
		\title{
			A Reinforcement Learning Framework for Safety-Aware Haptic Shared Control in Surgical Robotics
		}
		
		\author{\makebox[\linewidth][c]{Zihao~Zhang\,\orcidlink{0000-0002-2915-5503}, Karl~Handwerker\,\orcidlink{0009-0002-6088-4752}, Balint~Varga,~\IEEEmembership{Member,~IEEE}\,\orcidlink{0000-0002-4189-0105}}%
		\thanks{The authors are with the Institute of Control Systems, Karlsruhe Institute of Technology, 76131 Karlsruhe, Germany} \thanks{Corresponding author: Balint Varga, e-mail: balint.varga2@kit.edu.}}		
		
		\pagenumbering{gobble}

		\maketitle
		\thispagestyle{firstpage}% firstpage page style for first page
		\pagestyle{empty}

		\begin{abstract}
			Haptic shared control in surgical robotics combines human-side and robot-side forces at a shared end effector while preserving operator involvement. However, the effects of geometric safety choices on task execution and force coordination remain unclear. We model the interaction as a two-agent cooperative Markov game under centralized training and decentralized execution, \majorrev{comparing learned human control with a task-tuned fixed-impedance baseline.} This model-free multi-agent reinforcement learning framework supports joint policy adaptation without requiring a complete analytical interaction model. It compares MADDPG and rMAPPO with an epigraph-constrained approach under common geometric task definitions. In simulation, rMAPPO generally achieved the highest return and tracking performance, whereas the epigraph approach produced more conservative motions and lower geometric exposure, including no observed collision or dangerous-region exposure after geometry transfer. \majorrev{With real operator input, fixed-impedance-trained policies were faster and successful in all recorded trials. Policies trained with a learned partner exhibited greater command opposition, while the fixed epigraph condition produced the most aligned commands.} This work provides a benchmark for studying human-control representation and constraint-handling trade-offs in haptic shared control.
			
		\end{abstract}
		
		
		\section{Introduction}
		\label{sec:introduction}
		
		\IEEEPARstart{T}{eleoperated} surgical robots and shared control systems use force coupling to support precise and safe task execution while preserving the surgeon's authority under stringent clinical constraints~\cite{Li2023,Chen2024,Ebrahimi2021TBME,Patel2022}. Fully manual teleoperation preserves human authority but often imposes high workload, whereas full autonomy in surgery remains difficult to deploy widely because of safety, regulatory, and accountability concerns \cite{Brunke2021,Li2023,Chen2024,Yang2017,Lee2024}. Human-robot shared control therefore offers a clinically relevant middle ground between pure teleoperation and full autonomy \cite{Suresh2024,Tian2023TMRB,Varga2025Sensors}.
		
		In haptic shared control setting considered here, the human and the robot jointly influence a shared surgical instrument via combined force commands applied at a shared end effector. Such settings require accurate motion execution and strict control of tool-environment interaction for safe manipulation~\cite{Li2019,Music2020,Qin2025,Tian2023TMRB,Su2019RAL}.
		
		Force feedback can further improve operator awareness, motion quality, and force-related safety \cite{Enayati2016,Patel2022,Bergholz2023}. Since the human and the robot are tightly coupled on the same task, purely supervisory or leader-follower formulations are often insufficient. Instead, this setting is more naturally viewed as a cooperative sequential decision problem within the shared control paradigm.
		
		From this perspective, the interaction can be formulated as a multi-agent reinforcement learning (MARL) problem, in which the human-control and robot-control sides are modeled as coupled agents with a shared task objective.
		This model-free formulation permits joint policy optimization without a complete analytical model of the coupled behavior.
		However, MARL for this tightly coupled form of shared control remains only weakly explored. In surgical robotics, existing MARL work mainly addresses coordination among robotic agents \cite{Scheikl2021}, whereas continuous force coordination between a learned human control policy and a robot policy remains insufficiently studied.
		
		The choice of human control representation is another open question.
		Analytical formulations based on fixed impedance parameters or predefined cost functions assume essentially static behavior~\cite{Music2020,varga_toward_2024}, while data-driven methods trained offline do not jointly adapt the human and robot policies through interaction during training~\cite{Hu2023,Su2021TASE,Zheng2024ICRA}.
		Consequently, online bidirectional co-adaptation remains insufficiently captured. How robot policies trained with learned or fixed human control perform with real operators remains unclear.

		Safety handling in haptic shared control remains an open research problem. Penalty-based methods encode safety as a soft objective rather than an explicit constraint \cite{Brunke2021}, while shielding methods depend on sufficiently accurate models of the coupled human--robot dynamics \cite{Qin2025}. Epigraph-based MARL supports low-violation learning \cite{Zhang2025}, but its performance--safety trade-offs in haptic shared control remain insufficiently studied \cite{Gu2024,Fan2024}.
		
		To address these issues, we establish a framework for simulation and real operator evaluation of haptic shared control with safety-aware MARL under centralized training and decentralized execution.
		The framework compares learned and fixed human control conditions and evaluates penalty learning and the epigraph route under a common environment, geometric partition, and evaluation criteria. Frozen robot	policies are then tested with two real operators through a calibrated coupling between the device and simulation. 
		\newpage
		The main contributions of this paper are as follows:
		\begin{enumerate}
			
			\item We formulate the shared control problem as a cooperative Markov game, enabling force-level shared control without a complete analytical model of the coupled interaction.
			
			\item We compare a learned human control policy with a fixed impedance baseline, then evaluate the resulting frozen robot policies with
				real operator input through a coupling interface calibrated from the same two operators' motion data.
			
			\item We adapt an epigraph-constrained MARL formulation to this shared control setting for explicit geometric safety learning and compare it with MADDPG \cite{Lowe2017} and rMAPPO \cite{Yu2022MAPPO} under common evaluation criteria.
			
			\item We design a geometric reward and constraint design based on a region partition to compare penalty-based learning and the epigraph route. We evaluate all three routes across spherical, cubic, transferred, and rigid nonconvex geometries, complemented by real operator task and force allocation analyses.
 			
		\end{enumerate}

		\section{Related Work}
		
		\subsection{Shared Control in Surgical and Haptic Robotics}
		
		Shared control is an important paradigm in surgical robotics since it provides an intermediate level of assistance between fully manual teleoperation and higher levels of autonomy \cite{Li2023,Li2022,Chen2024}.
		
		Shared guidance or shared goal-level control keeps the human in the loop while the robot assists through virtual fixtures, path guidance, or safety intervention \cite{Tian2023TMRB,Su2019RAL,Varga2025Sensors,Iovene2025TBME}. For example, Tian \emph{et al.} \cite{Tian2023TMRB} and Iovene \emph{et al.} \cite{Iovene2025TBME} developed shared control frameworks for osteotomy that improve path accuracy and reduce operator burden, while Qin \emph{et al.} \cite{Qin2025} introduced a supervisory controller with force constraints for safer teleoperation. These approaches usually retain an asymmetric structure in which the robot guides, filters, or constrains the operator \cite{Li2023,Li2022,Iovene2025TBME,Tian2023TMRB}.
		
		In the tightly coupled \rev{haptic} shared control setting considered here, human and robot commands are generated in a common control space and combined at a shared end effector. Game theoretic approaches provide a formalism for this interaction \cite{Li2023,Li2022,Su2019RAL}. Music and Hirche \cite{Music2020} formulated a continuous-time differential game for haptic cooperative tracking and showed that game-theoretic modeling can improve cooperation while preserving intuitive interaction, assuming known human and robot cost functions. Li \emph{et al.} \cite{Li2019} modeled trajectory-level shared control as a nonzero-sum differential game and used reinforcement learning to approximate value functions online, enabling cooperative policies that balance task performance and interaction effort. These studies support viewing haptic shared control as a joint decision problem rather than a purely supervisory one.
		
		\rev{MARL has also been applied to high-level human-robot collaboration (HRC) problems, including multi-agent task scheduling and collaborative assembly or disassembly planning \cite{Yu2021,Xiao2023}. These studies demonstrate the usefulness of Markov game formulations for multi-agent coordination, but primarily address discrete task allocation and sequencing rather than continuous force coordination at a shared end effector.}

		\majorrev{Existing work rarely examines how a computational training controller affects a robot policy that is later paired with a real operator, while also comparing safety learning across varied geometry.} This motivates the cooperative Markov game formulation adopted in this work.
		
		\subsection{\majorrev{Human Control Representations}}
		
		\majorrev{A central question in shared control is how to represent the human contribution during controller development. Existing approaches include static control models, offline learning, joint adaptation, and hybrids of these strategies.}
		
		Static models describe the human contribution through impedance characteristics, predefined objectives, or analytical control principles \cite{Music2020,Li2023}. These formulations are interpretable and tractable, but assume fixed behavior and cannot express variation between operators or strategies.
		
		Offline learning methods infer task-relevant skills, assistance policies, latent intent, or user-related behavior from demonstrations through inverse reinforcement learning, imitation learning, or related techniques~\cite{Hu2023,Chen2024,Su2021TASE,Zheng2024ICRA,varga_toward_2024}.
		\rev{However, these approaches rely on offline-trained representations and do not jointly optimize human and robot policies through interaction during training.}
		
		\majorrev{Joint adaptation instead treats both control policies as adaptive decision makers that evolve through repeated computational interaction.}
		\majorrev{In this work, the learned human control policy and robot policy are
			initialized without demonstration pretraining and optimized jointly during
			simulation. The learned policy is a computational training partner, not a
			validated model of human neuromuscular behavior.}

		\rev{Hybrid paradigms combine demonstration based learning with subsequent
			adaptation to reduce exploration risk and improve data efficiency
			\cite{Shafti2020,Ou2023}. Co-GAIL, for example, learns a joint human-robot
			policy from human-human demonstrations and subsequently optimizes its two
			components during training \cite{Wang2023}. Such approaches depend on the
			availability and quality of demonstrations, which can be expensive to obtain
			in surgical settings \cite{Hu2023,Chen2024}.}

\begin{figure}[t]
			\centering
			\includegraphics[width=0.9\columnwidth]{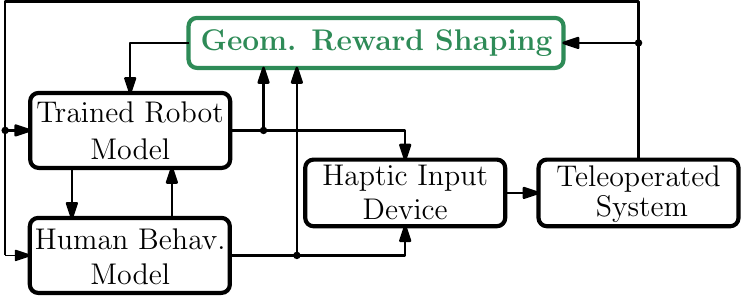}
			\caption{Haptic shared control with region partitioned reward and
					constraint design under learned and fixed human control conditions.}
			\label{fig:overview_framework}
		\end{figure}

		\majorrev{As illustrated in Fig.~\ref{fig:overview_framework}, we compare a
			fixed impedance baseline, for which only the robot policy is
			trained, with a learned condition in which both policies adapt jointly.}
		
		\majorrev{The resulting frozen robot policies are then paired with real
			operators to evaluate compatibility rather than fidelity to human behavior.}

		\subsection{Safety-Aware Learning for Surgical Collaboration}
		
		Safety is a primary concern in surgical robotics, where excessive interaction forces, unintended contacts, and workspace violations can compromise procedural quality~\cite{Qin2025,Gu2024,Fan2024,Ebrahimi2021TBME,Esfandiari2025TMRB,Pacchierotti2016TBME,Patel2022}. Penalty-based methods encode unsafe events directly in the reward and integrate easily with standard reinforcement learning algorithms, but they treat safety as a soft objective and therefore \rev{do not directly optimize an explicit safety constraint} \cite{Brunke2021}. This limitation is especially \rev{relevant} in surgical settings, where acceptable violation rates \rev{may be very low.}

		\majorrev{Runtime safety filters based on control barrier functions (CBFs) or
			quadratic programs (QPs) modify nominal commands online and can provide direct
			intervention when their model and feasibility assumptions hold
			\cite{Qin2025}. The epigraph route studied here instead incorporates the
			geometric constraint into policy learning, so the nominal policy is optimized
			under that constraint. These strategies are complementary: the present study
			does not establish superiority over CBF or QP filtering, whose application to
			tightly fused human-robot commands additionally requires a suitable interaction
			model \cite{Qin2025,Gu2024}.}
		
		Constrained optimization methods instead impose explicit safety
			constraints during learning through constrained Markov decision processes
			\cite{Brunke2021,Zhang2025}.
		Standard Lagrangian methods can become numerically unstable when the safety
		threshold approaches zero because the dual variable may grow excessively to
		enforce the constraint \cite{Zhang2025}.
		
		The epigraph formulation instead introduces an auxiliary variable
			and combines the safety and performance branches through a maximum operator.
			Its additive use of the auxiliary variable avoids scaling the policy gradient
			directly with a growing multiplier under zero or low violation objectives
			\cite{Zhang2025}.
		Although promising for safe multi-agent learning, this approach has not yet been studied in tightly coupled \rev{haptic} shared control, and systematic comparison between penalty-based and \majorrev{epigraph-constrained} safety learning \majorrev{across convex and nonconvex constraints} in this setting remains limited, see e.g.~\cite{Gu2024,Fan2024}.
		
		\section{Method}
		
		\subsection{Problem Formulation}
		
\begin{figure}[t]
			\centering
			\includegraphics[width=0.55\columnwidth]{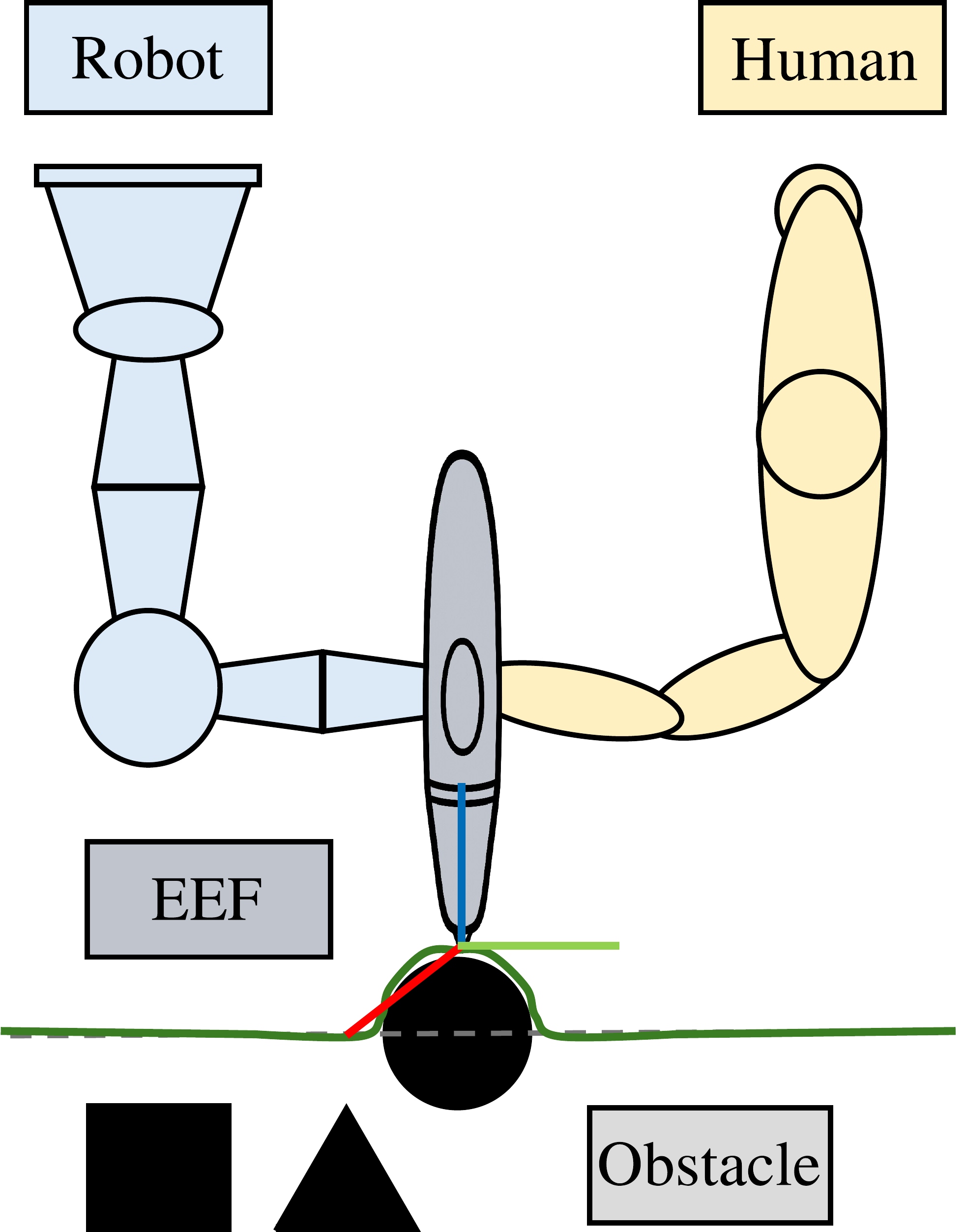}
			\caption{Schematic representation of the \rev{haptic shared control setting}.}
			\label{fig:overview}
		\end{figure}

		As illustrated in Fig.~\ref{fig:overview}, we formulate \rev{haptic} shared control as a two-agent cooperative Markov game under centralized training and decentralized execution.
		\majorrev{Let $H$ and $R$ denote the human and robot control inputs. During training, $R$ uses a learned policy $\pi^R$, whereas $H$ uses either a learned policy $\pi^H$ or fixed impedance. During real operator evaluation, the robot policy is frozen and the device input replaces the training controller.}
		\rev{Both sides receive the same six-dimensional observation
			$s_t\in\mathcal S$, where $\mathcal S$ is the shared observation space, and
			output normalized Cartesian force commands in $[-1,1]^3$.}
		The centralized critic input is
		\begin{equation}
			x_t=[s_t;s_t].
		\end{equation}
		\majorrev{The duplicated state preserves the joint critic interface; centralization is provided by the joint action, while each learned actor uses only $s_t$ at execution.}
		
		\majorrev{To express both human control conditions using the same learning
			equations, we define the normalized human command as}
		\begin{equation}
			\rev{
				\bar a_t^H=
				\begin{cases}
					\pi^H(s_t),
					& \text{learned condition},\\[1mm]
					\operatorname{clip}\!\left(
					f_{\mathrm{imp},t}/f_{\max},-1,1
					\right),
					& \text{fixed condition}.
				\end{cases}
			}
			\label{eq:human_command}
		\end{equation}
		\rev{The set of policies updated during training is}
		\begin{equation}
			\rev{
				\mathcal I_{\mathrm{learn}}=
				\begin{cases}
					\{H,R\}, & \text{learned condition},\\
					\{R\},   & \text{fixed condition}.
				\end{cases}
			}
			\label{eq:learned_agents}
		\end{equation}
		\majorrev{Thus, the fixed condition supplies the impedance command and
			updates only the robot actor; critic updates remain route-specific.}
		
		\majorrev{The shared end effector (EEF) receives the summed Cartesian
			commands:}
		\begin{equation}
			\rev{
				f_{\mathrm{total},t}=u_t^H+u_t^R,
				u_t^H=f_{\max}\bar a_t^H,
				u_t^R=f_{\max}a_t^R.
			}
			\label{eq:force_fusion}
		\end{equation}
		\majorrev{Here, $u_t^H$ and $u_t^R$ are the human and robot physical
			Cartesian force commands, $f_{\mathrm{total},t}$ is their sum,
			$a_t^R=\pi^R(s_t)$, and $f_{\max}=0.04\,\mathrm{N}$ maps normalized actions
			to per-axis force.}
		
		\subsection{\majorrev{Human Control Representations}}
		
		\majorrev{In the fixed condition, the human force is generated by the
			mass-free Cartesian impedance law}
		\begin{equation}
			\rev{
				\begin{aligned}
					\tilde f_{\mathrm{imp},t}
					&=
					K_p(p_t^{\mathrm{ref}}-p_t)
					+K_d(v_t^{\mathrm{ref}}-v_t),\\
					f_{\mathrm{imp},t}
					&=
					\operatorname{clip}\!\left(
					\tilde f_{\mathrm{imp},t},
					-f_{\mathrm{imp}}^{\max},
					f_{\mathrm{imp}}^{\max}
					\right),
				\end{aligned}
			}
			\label{eq:fixed_impedance}
		\end{equation}
		\majorrev{where, $\tilde f_{\mathrm{imp},t}$ and $f_{\mathrm{imp},t}$ are the
			unclipped and clipped impedance forces, and $(p_t,v_t)$ and
			$(p_t^{\mathrm{ref}},v_t^{\mathrm{ref}})$ are the current and reference EEF
			states. The proportional and damping gains are
			$K_p=0.6I_3\,\mathrm{N\,m^{-1}}$ and
			$K_d=0.075I_3\,\mathrm{N\,s\,m^{-1}}$; clipping is elementwise at
			$f_{\mathrm{imp}}^{\max}=0.03\,\mathrm{N}$ per axis, where $I_3$ is the
			identity matrix.}
		\majorrev{These gains were selected as a task-tuned baseline that produces a
			stable response and practical completion time. They were not identified as a
			physiological model from operator data. The fixed condition is therefore a
			controlled training reference for comparison with joint policy adaptation.}
		\rev{Using the trajectory quantities defined in
			Eq.~\eqref{eq:observation_geometry}, the reference position and velocity are}
		\begin{equation}
			\rev{
				\begin{aligned}
					p_t^{\mathrm{ref}}
					&=
					p_0+\operatorname{clip}(q_t+\ell,0,L)t_{\mathrm{dir}},\\
					v_t^{\mathrm{ref}}
					&=
					v_0\operatorname{clip}\!\left(
					\frac{L-q_t}{\ell},0,1
					\right)t_{\mathrm{dir}},
				\end{aligned}
			}
		\end{equation}
		\rev{where $\ell=0.04\,\mathrm{m}$ is the look-ahead distance and
			$v_0=0.02\,\mathrm{m\,s^{-1}}$ is the nominal reference speed, which is
			reduced near the goal.}

		\majorrev{For real operator evaluation, the first three Phantom Omni joints
			map to the simulated waist, shoulder, and elbow, while the distal joints are
			fixed. The mapped stylus position $p_t^{\mathrm{dev}}$ and its filtered finite
			difference velocity provide the device reference. The coupling gains were
			calibrated from motion trials collected from the same two operators who later
			performed the evaluation. For a trial of $T$ samples on motion axis $a$, the
			primary fit term was the position error}
		\begin{equation}
			\majorrev{
				\mathcal L_{\mathrm{pos}}=
				\frac{1}{T}\sum_{t=1}^{T}
				\left(p_{t,a}^{\mathrm{dev}}-p_{t,a}\right)^2,
			}
			\label{eq:device_calibration}
		\end{equation}
		\majorrev{where $p_t$ is the simulated EEF position. Candidate gains were
			screened by replaying the recorded device trajectories, with tracking error
			as the principal term and additional checks on lag, overshoot, command
			smoothness, and force saturation. Repeated online trials then established
			$K_H=1.7I_3\,\mathrm{N\,m^{-1}}$ and
			$D_H=0.32I_3\,\mathrm{N\,s\,m^{-1}}$, with a per-axis limit of
			$0.04\,\mathrm{N}$. This calibration from operator data applies only to the
			coupling between the device and simulation and is distinct from the fixed
			training baseline.}
		
		\majorrev{Because the device provides no direct operator force measurement,
			$u_t^H$ is the virtual impedance command reconstructed from the position and
			velocity tracking errors rather than a measured hand force. It replaces the
			training controller command in Eq.~\eqref{eq:force_fusion}.}
		
		\subsection{\majorrev{Penalty Learning Routes}}

		\majorrev{MADDPG and rMAPPO provide off- and on-policy penalty-based
			baselines under the same task interface and team reward; unlike the epigraph
			route, both maximize reward-based objectives.}
		
		\rev{For MADDPG,} the deterministic actor objective is
		\begin{equation}
			\rev{
				\max_{\theta_i}J_{\mathrm{actor}}^i
				=
				\mathbb E_{s_t\sim\mathcal D}
				\!\left[Q_{\psi_i}(x_t,A_t^\mu)\right],
			}
			\label{eq:maddpg_actor}
		\end{equation}
		\rev{for $i\in\mathcal I_{\mathrm{learn}}$, where}
		% \begin{equation}
			% A_t^{\mu}
			% =
			% [\mu_{\theta_H}(s_t),\,\mu_{\theta_R}(s_t)].
			% \end{equation}
		\begin{equation}
			\rev{
				A_t^\mu=
				\left[\bar a_t^H,\mu_{\theta_R}(s_t)\right].
			}
			\label{eq:maddpg_joint_action}
		\end{equation}
		
		\majorrev{Here, $A_t^\mu$ is the deterministic joint action,
			$\mu_{\theta_R}$ is the robot actor, and $Q_{\psi_i}$ is agent $i$'s
			centralized action-value critic; $\theta_i$ and $\psi_i$ are their parameters,
			and $\mathcal D$ is the replay buffer. Fixed-human training retains the
			impedance command in the robot critic input.}
		
		\rev{For rMAPPO, the recurrent actor maximizes}
		\begin{equation}
			\rev{
				\max_{\theta_i}J_{\mathrm{PPO}}^i
				=
				\mathbb E\!\left[
				L_t^{i,\mathrm{clip}}
				+c_{\mathrm{ent}}\mathcal H_t^i
				\right],
			}
			\label{eq:rmappo_objective}
		\end{equation}
		\rev{for $i\in\mathcal I_{\mathrm{learn}}$, where
			$\mathcal H_t^i$ is the policy entropy and $c_{\mathrm{ent}}$ is its
			coefficient.}
		\begin{equation}
			\rev{
				L_t^{i,\mathrm{clip}}
				=
				\min\!\left\{
				\rho_t^i\hat A_t^i,
				\operatorname{clip}(\rho_t^i,1-\epsilon,1+\epsilon)\hat A_t^i
				\right\}.
			}
			\label{eq:rmappo_clip}
		\end{equation}
		\majorrev{Here, $\hat A_t^i$, $\rho_t^i$, and $\epsilon$ denote the advantage,
			current-to-rollout likelihood ratio, and clipping threshold; separately
			parameterized recurrent value critics form the advantages.}
		
		\subsection{\majorrev{Epigraph Constrained} Learning Route}

		\majorrev{The epigraph route minimizes the performance cost
			$J_{\mathrm{perf}}=-J_{\mathrm{task}}$, where $J_{\mathrm{task}}$ is the
			expected team task return, subject to the signed geometric safety signal
			$h_t$ in Eq.~\eqref{eq:setup_signed_constraint}. Separate recurrent safety
			critics represent the same physical constraint for $H$ and $R$; with
			time-varying epigraph variable $z_t$, their maximum-type values are}
		\begin{equation}
			\rev{
				\begin{aligned}
					J_{\mathrm{safety}}^i(s_t,z_t)
					&=V_h^i(s_t,z_t),\\
					V_h^i(s_t,z_t)
					&=\mathbb E_\pi
					\!\left[\max\{h_t,G_t^i\}\right],\\
					G_t^i
					&=(1-\gamma)h_t
					+\gamma V_h^i(s_{t+1},z_{t+1}).
				\end{aligned}
			}
			\label{eq:safety_objective}
		\end{equation}
		\majorrev{where, for $i\in\{H,R\}$, $J_{\mathrm{safety}}^i$ is the safety
			objective, $V_h^i$ its maximum-type value estimate, $\mathbb E_\pi$ the
			joint-policy expectation, $\gamma$ the discount factor, and $G_t^i$ the
			one-step bootstrap target. Training uses truncated $\lambda$-returns, where
			$\lambda$ controls the return horizon; the constrained problem is
			\cite{Zhang2025}}
		\begin{equation}
			\rev{
				\begin{aligned}
					&\min_{z,\{\pi^j\}_{j\in\mathcal I_{\mathrm{learn}}}}
					\;z\\
					&\text{s.t.}\;\;
					\max\!\left\{
					J_{\mathrm{safety}}^i,
					J_{\mathrm{perf}}-z
					\right\}
					\le0.
				\end{aligned}
			}
			\label{eq:epigraph_problem}
		\end{equation}
		\majorrev{for each $i\in\{H,R\}$, while only actors in
			$\mathcal I_{\mathrm{learn}}$ are updated. One centralized performance
			critic, two safety critics, and $z$-conditioned recurrent networks yield}
		\begin{equation}
			\rev{
				V_t^{\mathrm{epi},i}
				=
				\max\!\left\{
				V_t^{\mathrm{safety},i},
				V_t^{\mathrm{perf}}-z_t
				\right\},
			}
		\end{equation}
		\majorrev{where the terms are the safety, performance, and combined value
			estimates. Actors minimize a clipped upper surrogate of the epigraph-cost
			advantage; the fixed condition still updates the performance and both safety
			critics.}
		
		\rev{The team task reward and the epigraph transition are}
		\begin{equation}
			\rev{
				\begin{aligned}
					r_{\mathrm{team},t}
					&=
					\frac{r_{\mathrm{task},t}^H+r_{\mathrm{task},t}^R}{2},\\
					z_{t+1}
					&=
					\Pi_{[z_{\min},z_{\max}]}
					\left(
					\frac{z_t+r_{\mathrm{team},t}}{\gamma}
					\right),
				\end{aligned}
			}
			\label{eq:team_task_reward}
		\end{equation}
		\majorrev{where, $r_{\mathrm{team},t}$ is a reward, and
			$\Pi_{[z_{\min},z_{\max}]}$ projects onto the admissible epigraph interval.}
		\majorrev{The terms $r_{\mathrm{task},t}^H$ and
			$r_{\mathrm{task},t}^R$ are the human and robot task reward
			components, and $z_{\min}$ and $z_{\max}$ are the interval bounds.}
		
		\majorrev{Both recurrent routes use one 256-unit GRU, a one-layer head, 256-step rollouts, and length-16 sequences. The epigraph route additionally embeds $z$ in six dimensions through two fully connected layers. With a fixed human, only the robot actor is updated; rMAPPO updates its robot value critic $V_R$, whereas the epigraph route updates $V_{\mathrm{perf}}$, $V_h^H$, and $V_h^R$.} \majorrev{At execution, each safety critic returns the smallest $z$ satisfying the target bound $h_{\mathrm{tgt}}$, and the larger value is used:}
		\begin{equation}
			\rev{
				\begin{aligned}
					z_i^\star
					&=\min\{z:V_h^i(s_t,z)\le h_{\mathrm{tgt}}\},\\
					z
					&=\max_{i\in\{H,R\}}z_i^\star.
				\end{aligned}
			}
		\end{equation}
		\majorrev{Binary search spans $[z_{\min},z_{\max}]=[-450,300]$, with
			$h_{\mathrm{tgt}}=-0.05$ providing a margin below zero under the approximate
			non-increasing relation between $V_h^i(s,z)$ and $z$.}

		% \section{Novel Reward Design Method}
		\section{\rev{Experimental Setup}}
		\label{sec:setup}
		
		\subsection{Task and Environment}
		
\begin{figure}[t]
			\centering
			\subfloat[\rev{Spherical obstacle}]{
				\includegraphics[width=0.47\columnwidth]{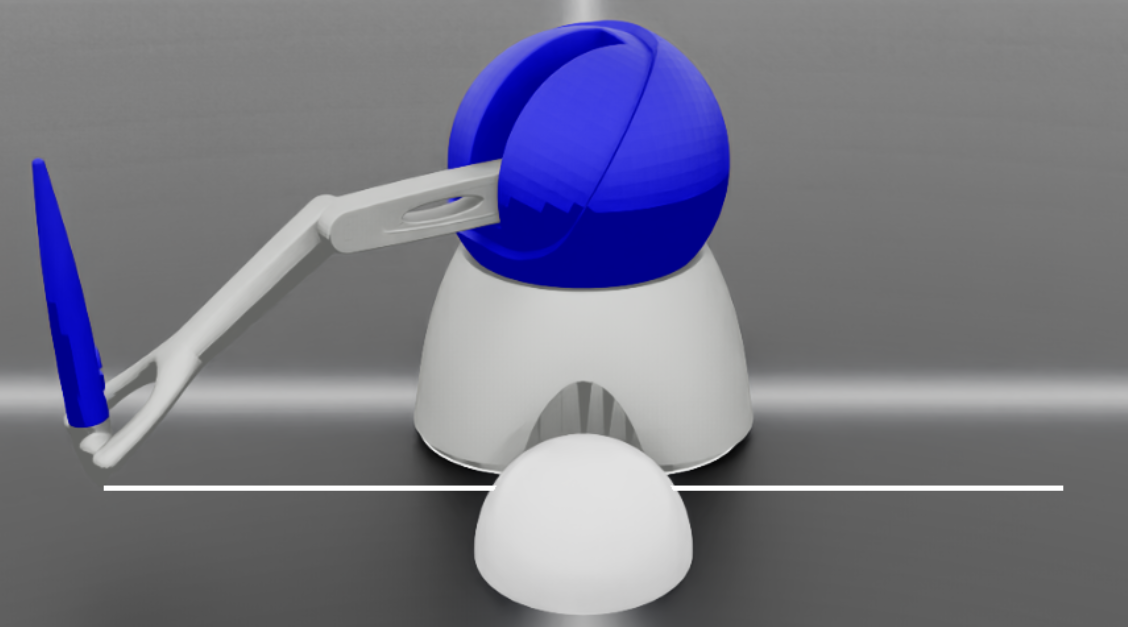}
			}
			\hfill
			\subfloat[\rev{Cubic obstacle}]{
				\includegraphics[width=0.47\columnwidth]{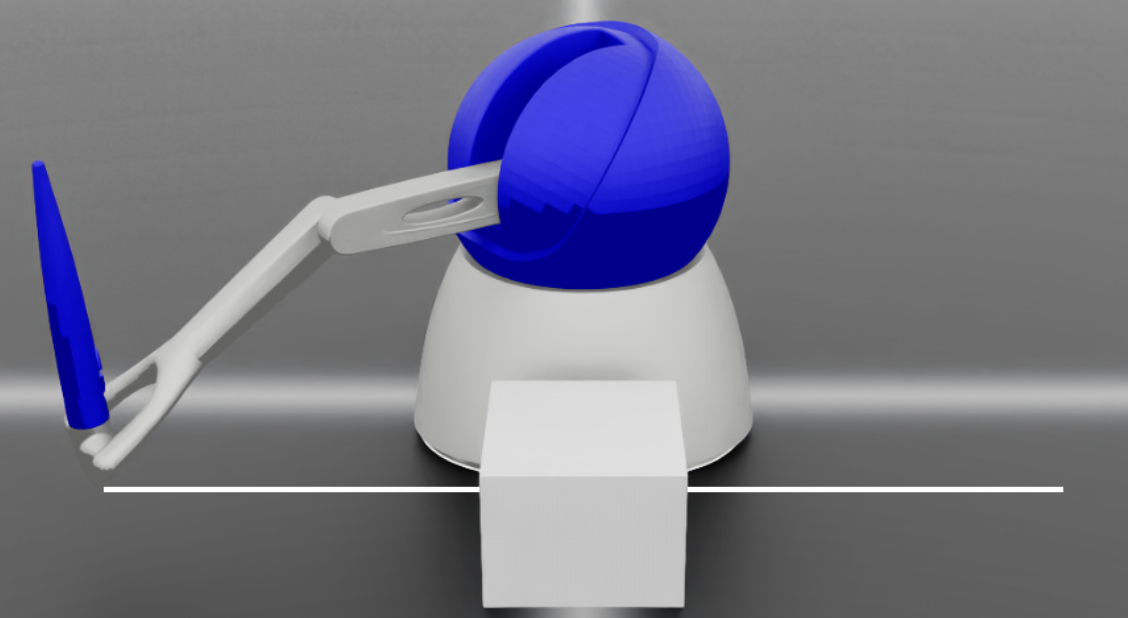}
			}
			\\[0.55em]
			\subfloat[\majorrev{Rigid nonconvex obstacle}]{
				\includegraphics[width=0.47\columnwidth]{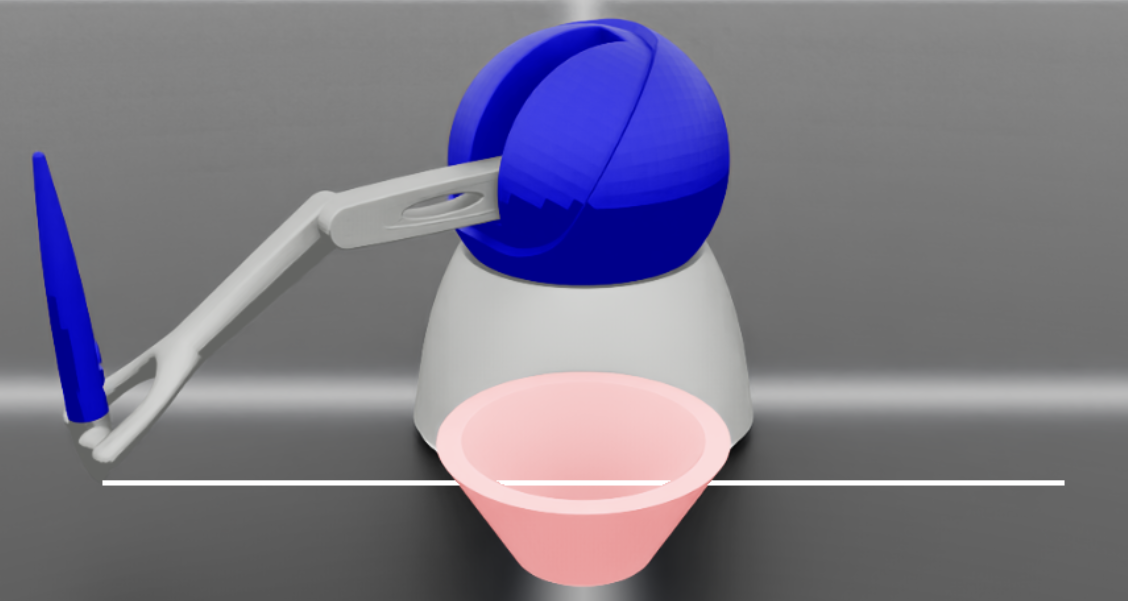}
			}
			\caption{\majorrev{Task scenarios with spherical, cubic, and rigid
					nonconvex obstacles.}}
			\label{fig:scenarios}
			\vspace{-0.8em}
		\end{figure}

		\rev{The experiments are conducted in Isaac Lab on an abstracted haptic
			shared control task.}
		\rev{The task abstracts pre-contact surgical navigation: the two sides apply
			forces to a shared EEF that follows a reference trajectory, generates its own
			avoidance motion around a rigid obstacle, and then rejoins the path. No
			obstacle-avoidance trajectory is prescribed.}
		\majorrev{Human control is provided by either a learned policy or fixed
			impedance, and replaceable obstacle geometries support the evaluation settings
			described below.}
		\majorrev{The evaluated constraints are spherical, cubic, and rigid
			nonconvex; the last is a concave, perforated conical shell whose nominal
			path crosses two separated walls. All use the same six-dimensional
			observation, force limits, region partition, reward, and safety definitions.}
		\majorrev{The three obstacle scenarios are illustrated in
			Fig.~\ref{fig:scenarios}.}
		
		At each step, the environment provides obstacle-related geometric feedback, \rev{and both sides receive the same observation}
		\begin{equation}
			s_t=[\mathrm{dev}_t;\,v_t^\top;\,d_t;\,
			\mathrm{prog}_t]^\top\in\mathbb R^6,
		\end{equation}
		\rev{Here, $v_t$ is the EEF velocity and $d_t$ is the unsigned distance to the obstacle surface. Let $p_0$, $p_g$, and $p_t$ denote the start, goal, and current EEF positions, respectively. The trajectory quantities used in the observation and fixed impedance controller are}
		\begin{equation}
			\rev{
				\begin{aligned}
					L&=\lVert p_g-p_0\rVert,\qquad
					t_{\mathrm{dir}}=(p_g-p_0)/L,\\
					q_t&=\operatorname{clip}\!\left(
					(p_t-p_0)^\top t_{\mathrm{dir}},0,L\right),\\
					\mathrm{dev}_t&=\lVert p_t-p_0-q_tt_{\mathrm{dir}}\rVert_2,\\
					\mathrm{prog}_t&=(p_t-p_0)^\top t_{\mathrm{dir}}/L.
				\end{aligned}
			}
			\label{eq:observation_geometry}
		\end{equation}
		\rev{Here, $L$ is the straight-line path length, $t_{\mathrm{dir}}$ is its
			unit direction, $q_t$ is the clipped along-path distance,
			$\mathrm{dev}_t$ is the perpendicular trajectory deviation, and
			$\mathrm{prog}_t$ is the unclipped normalized progress. The surface normal
			and binary overlap indicator are used only for reward and safety
			construction, with $o_t=1$ denoting geometric overlap. We use $s_t$ for the
			shared observation rather than the full simulator state.}
		\rev{Episodes terminate upon successful task completion, defined as the EEF
			reaching within $0.01\,\mathrm{m}$ of the goal, or at the time limit.}
		\rev{Safety-related events do not cause early termination. The environment
			parameters are listed in Table~\ref{tab:sim_params}.}
		
		\begin{table}[t]
			\caption{Key Environment Parameters}
			\label{tab:sim_params}
			\centering
			\begin{tabular}{@{}lc@{}}
				\hline
				Parameter & Value \\
				\hline
				\rev{Learned-policy force limit $f_{\max}$}
				& \rev{$0.04\,\mathrm{N/axis}$} \\
				\rev{Fixed impedance limit $f_{\mathrm{imp}}^{\max}$}
				& \rev{$0.03\,\mathrm{N/axis}$} \\
				Simulation frequency & $120\,\mathrm{Hz}$ \\
				Control frequency & $60\,\mathrm{Hz}$ \\
				Episode horizon & $1200\,\mathrm{steps}$ \\
				Completion radius & $0.01\,\mathrm{m}$ \\
				Sphere radius & $0.05\,\mathrm{m}$ \\
				Cube side length & $0.08\,\mathrm{m}$ \\
				Cube shift (generalization) & $0.02\,\mathrm{m}$ \\
				\majorrev{Rigid nonconvex extent}
				& \majorrev{$0.12\times0.12\times0.052\,\mathrm{m}$} \\
				\hline
			\end{tabular}
			\compacttablebottom
		\end{table}
		
		\subsection{Reward and Safety Design}
		
		The reward and safety design is based on a unified geometric framework
		defined on a shared region partition. For the penalty-based methods, signed
		reward-shaping terms encode task progress and safety-related penalties within
		a single team reward.
		\rev{For the epigraph method, we retain the same geometric partition and task
			semantics, while representing geometric safety through the signed
			maximum-type constraint signal $h_t$ defined below.} Thus, the geometric structure is shared across methods, \rev{whereas the constrained safety branch is separated from the task reward and is not constructed as a cumulative nonnegative violation cost.}
		
		To represent trajectory tracking, obstacle-surface following, \rev{danger avoidance}, and trajectory rejoining, the task space is partitioned into four non-overlapping regions
		\rev{$\mathcal Z(s_t)\in\{A,B,C,D\}$, where $\sum_{Z\in\{A,B,C,D\}}\mathbf 1_Z(s_t)=1$. Let $d_C=0.0075\,\mathrm{m}$ and $d_A=0.015\,\mathrm{m}$ denote the inner and outer surface-distance thresholds, respectively. Zone~A is active when $d_t\ge d_A$, and Zone~C is active when $d_t\le d_C$. Within the surface band $d_C<d_t<d_A$, Zone~D is activated when the rejoining condition described below holds; the remainder is Zone~B.} Zone~C is divided into a danger subregion ($o_t=0$), where proximity and inward-velocity penalties apply, and an overlap subregion ($o_t=1$), where a dedicated collision penalty replaces \rev{those terms}.
		
		\rev{The implemented Zone~D detector uses the unit goal direction $g_t=(p_g-p_t)/\lVert p_g-p_t\rVert$, trajectory direction $t_{\mathrm{dir}}$, and the unit direction $n_t$ from the EEF toward the obstacle. A surface-band state is classified as Zone~D after} 
		\begin{equation}
			\rev{
				\langle g_t,t_{\mathrm{dir}}\rangle\ge0.90,
				\langle n_t,t_{\mathrm{dir}}\rangle\le-0.866
			}
			\label{eq:zone_d_rule}
		\end{equation}
		\rev{holds for ten consecutive control steps. Zone~B is therefore the surface-band mask excluding Zone~D.}
		
		\rev{Let $t_{\mathrm{surf},t}$ be the normalized projection of $t_{\mathrm{dir}}$ onto the obstacle tangent plane.} The projected velocities are
		\begin{equation}
			\begin{aligned}
				v_{\mathrm{along},t}&=\langle v_t,t_{\mathrm{dir}}\rangle,\\
				v_{\mathrm{norm},t}&=\langle v_t,n_t\rangle,\\
				v_{\mathrm{surf},t}&=\langle v_t,t_{\mathrm{surf},t}\rangle.
			\end{aligned}
		\end{equation}
		\rev{Here, $v_{\mathrm{along},t}$ measures motion along the reference path
			and $v_{\mathrm{surf},t}$ measures tangential bypass motion. Because $n_t$
			points toward the obstacle, $v_{\mathrm{norm},t}>0$ corresponds to inward
			motion and $v_{\mathrm{norm},t}<0$ to outward motion.}
		
		\subsubsection{Region-Specific Shaping}
		
		In Zones~A and D, the shaping emphasizes progress and tracking. The progress
		reward is
		\begin{equation}
			\rev{r_{\mathrm{prog},t}=
				\begin{cases}
					-\mathcal R(\mathrm{prog}_t),
					& \mathrm{prog}_t>1,\\
					\mathcal R(\mathrm{prog}_t),
					& \mathrm{prog}_t\le1,\;
					v_{\mathrm{along},t}>\epsilon_v,\\
					-\mathcal R(\mathrm{prog}_t),
					& \mathrm{prog}_t\le1,\;
					v_{\mathrm{along},t}<-\epsilon_v,\\
					0,
					& \mathrm{prog}_t\le1,\;
					|v_{\mathrm{along},t}|\le\epsilon_v,
			\end{cases}}
		\end{equation}
		where \rev{$\mathcal R(p)=r_{\min}+(r_{\max}-r_{\min})p$ scales the reward magnitude linearly with progress, and $\epsilon_v>0$ is a small velocity
			deadband. Tracking quality is enforced via}
		\begin{equation}
			\rev{
				r_{\mathrm{dev},t}
				=-\phi_{\mathrm{dev}}(\mathrm{dev}_t),
			}
		\end{equation}
		\rev{where $\phi_{\mathrm{dev}}$ is a nondecreasing clipped function that is
			approximately linear for small deviations and saturates for large ones.}
		
		In Zone~B, the rewards focus on safe \rev{boundary following}:
		\begin{equation}
			\begin{aligned}
				r_{\mathrm{gap},t}
				&=-k_{\mathrm{gap}}(d_t-d^*)^2,\\
				r_{\mathrm{surf},t}
				&=k_{\mathrm{surf}}v_{\mathrm{surf},t},\\
				r_{\mathrm{inw},B,t}
				&=-k_{\mathrm{inw}}[v_{\mathrm{norm},t}]_+,
			\end{aligned}
		\end{equation}
		where $d^*$ is the desired safety clearance and $k_{\mathrm{gap}}>0$ is the gap penalty stiffness. \rev{We set $d^*=0.010\,\mathrm{m}$.} \rev{The coefficient} $k_{\mathrm{surf}}>0$ scales the reward for tangential bypass motion, and $k_{\mathrm{inw}}>0$ penalizes inward motion;
		$[\cdot]_+$ denotes the positive part.
		
		In Zone~C, \rev{danger and overlap} are handled separately:
		\begin{equation}
			\begin{aligned}
				r_{\mathrm{inw},C,t}
				&=-k_{\mathrm{inw}}[v_{\mathrm{norm},t}]_+,\\
				r_{\mathrm{off},t}
				&=R_{\mathrm{off}},\\
				r_{\mathrm{overlap},t}
				&=R_{\mathrm{overlap}},
			\end{aligned}
		\end{equation}
		\rev{where $R_{\mathrm{off}}<0$ and
			$R_{\mathrm{overlap}}<R_{\mathrm{off}}$ are the danger-region and overlap
			penalties, respectively.} In the danger subregion,
		\rev{$r_{\mathrm{off},t}$} and \rev{$r_{\mathrm{inw},C,t}$} are active simultaneously. \rev{Once overlap is detected, only $r_{\mathrm{overlap},t}$ applies.} \rev{In Zone~D,}
		\begin{equation}
			r_{\mathrm{inw},D,t}
			=-k_{\mathrm{inw}}v_{\mathrm{norm},t},
		\end{equation}
		\rev{The Zone~D term rewards outward motion and penalizes inward motion
			during rejoining.}
		
		\begin{table}[!t]
			\caption{\rev{Task Terms Under the Unified Region Partition}}
			\label{tab:reward_assignment}
			\centering
			\begin{tabularx}{\columnwidth}{@{}llXcc@{}}
				\hline
				\noalign{\vskip 3pt}
				\makecell[l]{Region} & \makecell[l]{Symbol} & \makecell[l]{Description}
				& \makecell[c]{Team\\reward}
				& \makecell[c]{Epi.\\task}\\
				\noalign{\vskip 3pt}
				\hline
				Zone A & $r_{\mathrm{prog}}$ & Progress reward
				& \checkmark & \checkmark\\
				Zone A & $r_{\mathrm{dev}}$ & Deviation penalty
				& \checkmark & \checkmark\\
				Zone B & $r_{\mathrm{gap}}$ & Gap penalty
				& \checkmark & \checkmark\\
				Zone B & $r_{\mathrm{surf}}$ & Surface-tangential reward
				& \checkmark & \checkmark\\
				Zone B & $r_{\mathrm{inw},B}$ & Inward-motion penalty
				& \checkmark & $\times$\\
				Zone C & $r_{\mathrm{off}}$ & Danger-region penalty
				& \checkmark & $\times$\\
				Zone C & $r_{\mathrm{inw},C}$ & Inward-motion penalty
				& \checkmark & $\times$\\
				Zone C & $r_{\mathrm{overlap}}$ & Overlap penalty
				& \checkmark & $\times$\\
				Zone D & $r_{\mathrm{prog}}$ & Progress reward
				& \checkmark & \checkmark\\
				Zone D & $r_{\mathrm{dev}}$ & Deviation penalty
				& \checkmark & \checkmark\\
				Zone D & $r_{\mathrm{inw},D}$ & Normal-velocity regulation
				& \checkmark & $\times$\\
				Global & $r_{\mathrm{comp}}$ & Completion reward
				& \checkmark & \checkmark\\
				Global & $r_{\mathrm{pot}}$ & Goal-potential reward
				& \checkmark & \checkmark\\
				Global & $r_{\mathrm{force}}^i$ & \rev{Individual effort}
				& \checkmark & \checkmark\\
				Global & $r_{\mathrm{aware}}^i$ & \rev{Partner effort}
				& \checkmark & \checkmark\\
				\hline
			\end{tabularx}
			\compacttablebottom
		\end{table}
		
		\rev{The epigraph safety branch does not sum the excluded reward penalties. Let $d_0=5\times10^{-4}\,\mathrm{m}$ be the clearance threshold, $c_{\mathrm{scale}}=0.015\,\mathrm{m}$ the normalization length, and $c_t=d_t-d_0$ the signed clearance margin. The shared signed constraint is}
		\begin{equation}
			\rev{h_t=
				\begin{cases}
					-\min(c_t/c_{\mathrm{scale}},1),
					& \text{safe},\\
					\min(0.5+\max(-c_t,0)/d_0,1),
					& \text{unsafe},
				\end{cases}
			}
			\label{eq:setup_signed_constraint}
		\end{equation}
		\rev{where unsafe means $o_t=1$ or $c_t<0$. Hence $h_t\le0$ in safe states
			and $h_t>0$ under violation. The constraint boundary $d_0$ is distinct from
			the broader Zone~C threshold $d_C$: the latter is used to report proximity
			exposure, whereas $h_t>0$ denotes an epigraph constraint violation. Its
			maximum-type value is defined in Eq.~\eqref{eq:safety_objective}.}
		
		\subsubsection{Global Terms and Aggregation}
		\rev{The completion term $r_{\mathrm{comp},t}$ equals $R_{\mathrm{comp}}>0$ when
			$d_{\mathrm{goal},t}=\lVert p_t-p_g\rVert$ falls below the completion radius, and is zero otherwise.} \rev{The goal-proximity reward is}
		\begin{equation}
			\rev{r_{\mathrm{pot},t}
				=
				\sqrt{
					\operatorname{clip}\!\left(
					1-\frac{d_{\mathrm{goal},t}}{d_{\mathrm{goal},0}},
					0,1
					\right)
				},}
		\end{equation}
		where $d_{\mathrm{goal},0}$ is the initial distance to the goal. \rev{The effort terms are}
		\begin{equation}
			\rev{\begin{aligned}
					r_{\mathrm{force},t}^i
					&=-\lVert u_t^i\rVert_2,\\
					r_{\mathrm{aware},t}^i
					&=-\lVert u_t^{-i}\rVert_2,
			\end{aligned}}
		\end{equation}
		\rev{where, $-i$ denotes the other side. The first term penalizes individual
			effort, while the second accounts for the partner contribution, thereby
			discouraging excessive total effort and excessive reliance on either force
			contribution.}
		
		\begin{figure}[t]
			\centering
			\includegraphics[width=0.9\columnwidth]{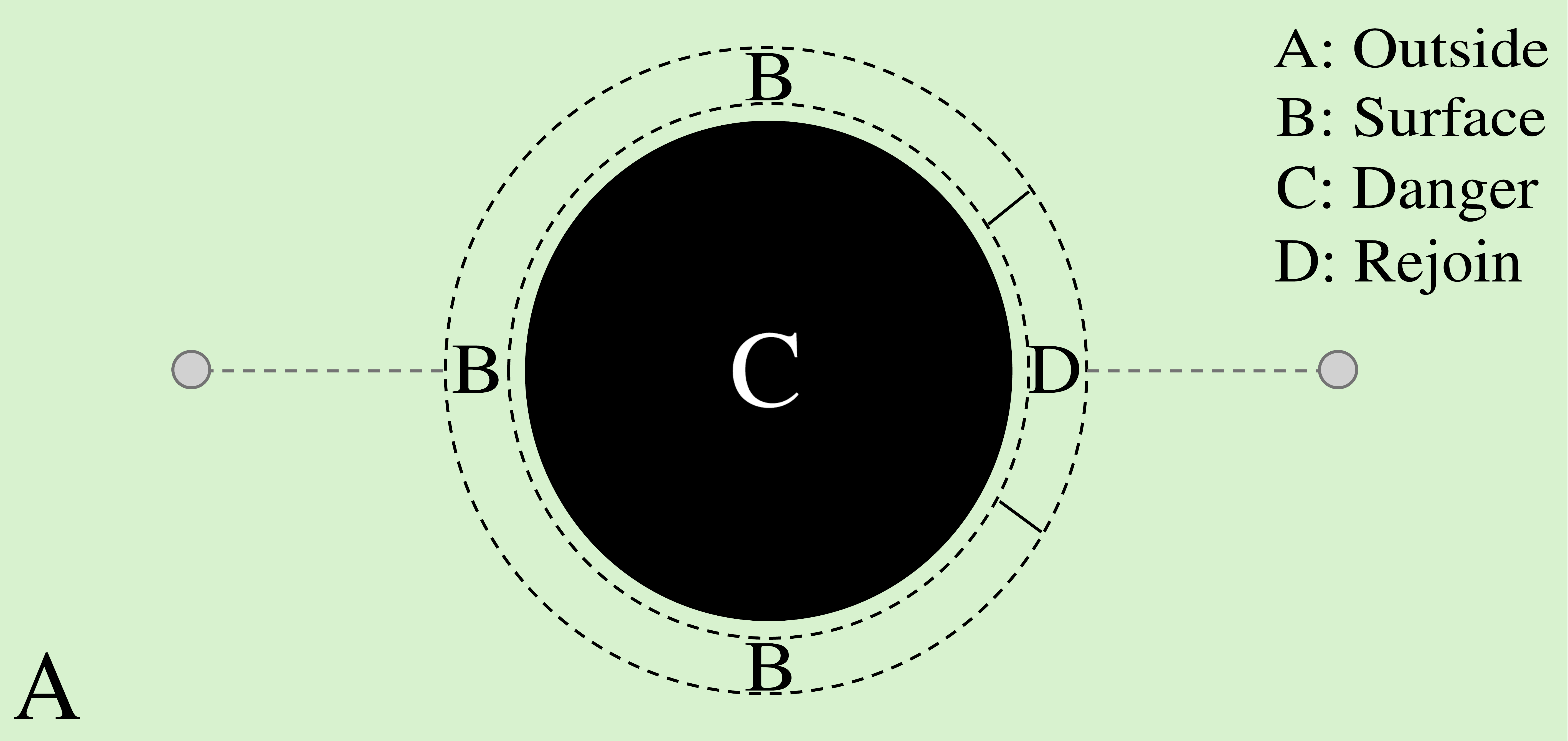}
			\caption{Four-zone geometric partition around the obstacle.}
			\label{fig:zone_partition}
		\end{figure}
		\rev{For MADDPG and rMAPPO, the regional and global terms are aggregated as}
		\begin{equation}
			\rev{
				\begin{aligned}
					r_t^i
					&=\sum_{Z\in\{A,B,C,D\}}r_{Z,t}^i
					+r_{\mathrm{global},t}^i,\\
					r_{\mathrm{team},t}
					&=\tfrac12(r_t^H+r_t^R).
				\end{aligned}
			}
		\end{equation}
		\rev{Here, $r_{Z,t}^i$ is the weighted regional reward, which is zero outside
			region $Z$, and $r_{\mathrm{global},t}^i$ is the weighted sum of global
			terms. Thus, $r_{\mathrm{team},t}$ denotes the mean complete shaped reward
			for the penalty-based routes; in the epigraph route,
			Eq.~\eqref{eq:team_task_reward} instead uses the mean task-only reward.}
		\rev{For epigraph, the task-related primitives are aggregated as}
		\begin{equation}
			\rev{
				r_{\mathrm{task},t}^i
				=
				\sum_{(Z,k)\in I_{\mathrm{task}}}
				w_{Z,k}^ir_{Z,k,t}
				+r_{\mathrm{global},t}^{i,\mathrm{task}},
			}
		\end{equation}
		\rev{where $Z$ indexes the region, $k$ indexes the reward primitive,
			$w_{Z,k}^i\ge0$ is its coefficient, and $I_{\mathrm{task}}$ is the set of
			region-primitive pairs assigned to the epigraph task branch in
			Table~\ref{tab:reward_assignment}. The term
			$r_{\mathrm{global},t}^{i,\mathrm{task}}$ collects the global primitives
			assigned to that branch in the same table.}
		\rev{The safety branch instead uses the shared $h_t$ in
			Eq.~\eqref{eq:setup_signed_constraint}.}
		\rev{The principal dense per-step reward terms are scaled to magnitudes of
			approximately $0.1$ to $0.4$ to limit value growth and premature action
			saturation.}

		\subsection{Evaluation Protocol}
		
		\majorrev{Sphere experiments cover both human control conditions; cube experiments use the learned condition, including direct transfer of the sphere-trained policies. These settings use four seeds, with frozen actors and deterministic evaluation.}
		\majorrev{For the rigid nonconvex geometry, three frozen policies per method, one from each seed, are evaluated under the fixed impedance condition. Each policy is executed deterministically for one episode, and the results are summarized across seeds.}
		
		\rev{For descriptive cross-route comparison, the normalized composite score is}
		\begin{equation}
			\bar R=\frac{10^3R_{\mathrm{ep}}}{L_{\mathrm{task}}},
			\label{eq:normalized_composite_score}
		\end{equation}
		\majorrev{where, $R_{\mathrm{ep}}$ and $L_{\mathrm{task}}$ are the episode
			return and executed steps. MADDPG and rMAPPO use the mean reward of the two control inputs;
			the epigraph route uses
			$\frac12\sum_{i\in\{H,R\}}(r_{\mathrm{task},t}^i-c_{\mathrm{eval},t}^i)$,
			where
			$c_{\mathrm{eval},t}^i=[-r_{\mathrm{risk},t}^i]_+$ aggregates excluded safety
			shaping terms. Checkpoints maximize mean $\bar R$ across seeds at common
			milestones for the sphere/cube studies. This post-hoc score has no independent
			validation set and is interpreted only with the geometric metrics.} \majorrev{We report $P_{\mathrm{fin}}$, $\mathrm{Steps}$, $\bar R$, $\rho_{\mathrm{track}}$, $\rho_C$, and $\rho_{\mathrm{col}}$. $P_{\mathrm{fin}}$ is final unclipped progress; the three $\rho$ terms are the fractions of steps with $\mathrm{dev}_t<0.01\,\mathrm{m}$, $d_t\le d_C$, and $o_t=1$, respectively. Timeouts use the full horizon.
			Simulation results are averaged within seed and then across seeds; all
			training panels show mean $\pm$ sample standard deviation. Force profiles
			describe coordination between the two inputs.}
		\majorrev{For the nonconvex task, Success is the fraction of runs entering
			the $0.010\,\mathrm{m}$ goal radius, and trajectory RMSE is
			$\sqrt{L_{\mathrm{task}}^{-1}\sum_t\mathrm{dev}_t^2}$.}

		\majorrev{The real operator feasibility pilot evaluates six frozen robot policies spanning the three routes and two training conditions. Device input replaces the training controller. The two operators whose motion trials were used for coupling calibration each complete five spherical episodes per condition, giving 60 episodes. Each operator completed five valid spherical-task episodes per condition, yielding 60 episodes. Because the study involved only two operators, the results are interpreted as feasibility evidence rather than population-level findings.}
		
\begin{figure}[!t]
			\centering
			\includegraphics[width=0.99\columnwidth]{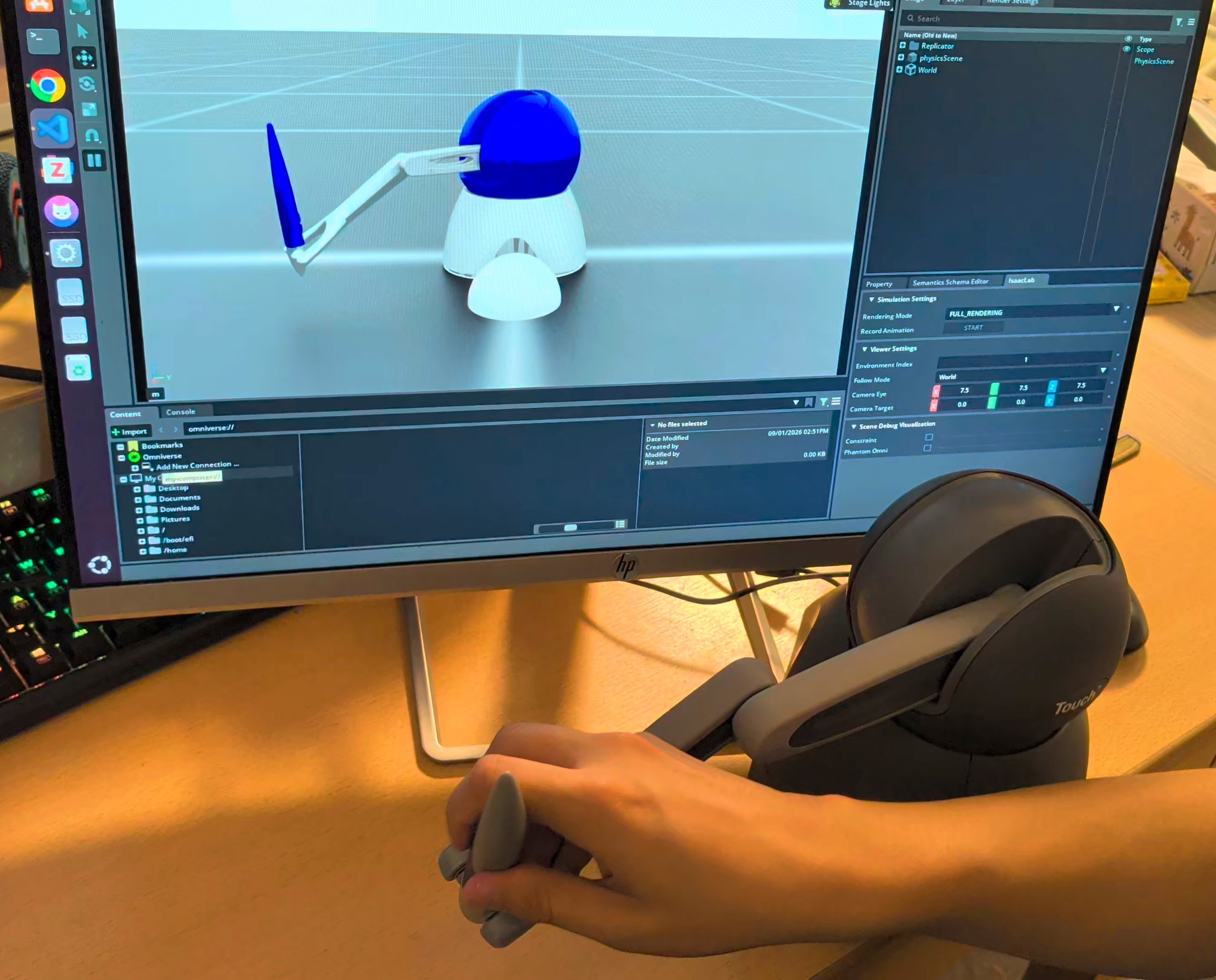}
			\caption{Real operator setup coupling the Phantom Omni input device
					to the Isaac Sim environment.}
			\label{fig:hardware_setup}
			\vspace{-0.6em}
		\end{figure}

		\majorrev{The study protocol was approved by the ethics committee (No. 596). Both participants gave voluntary written informed consent after being advised of their right to withdraw at any time without penalty. All data were fully anonymized prior to analysis.
			Each episode begins after alignment within $0.010\,\mathrm{m}$ of the start and a $0.5\,\mathrm{s}$ stationary hold; the same goal radius and 1200-step horizon apply. The setup is shown in Fig.~\ref{fig:hardware_setup}.}

		\majorrev{The device-to-EEF tracking RMSE is the root mean square distance between the mapped Phantom reference and simulated EEF, excluding the terminal reset sample. With force-time integrals $E_i=\int\lVert u^i(t)\rVert_2dt$ and $E_{\mathrm{comb}}=\int\lVert u^H(t)+u^R(t)\rVert_2dt$, human share is $\eta_H=E_H/(E_H+E_R)$ and cancellation is $\chi=1-E_{\mathrm{comb}}/(E_H+E_R)$. The mean force cosine and opposing ratio are computed where both force norms exceed $10^{-3}\,\mathrm{N}$. Metrics are calculated per episode and reported as descriptive mean $\pm$ sample standard deviation over ten episodes per condition, or three seeds for the nonconvex task. Bold values indicate the best result within each comparison group.}

		\section{Results}
		\label{sec:results}
		
		\definecolor{gridlight}{gray}{0.82}
		\definecolor{gridmajor}{gray}{0.68}
		\definecolor{axisgray}{gray}{0.55}
		
		\begin{figure*}[!t]
			\centering
			\subfloat[\rev{Learned: sphere}]{%
				\includegraphics[width=0.475\textwidth]
				{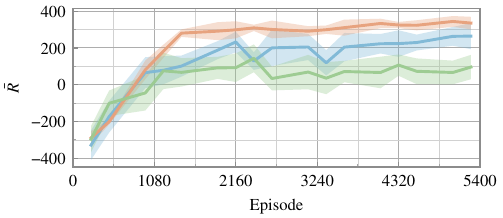}%
			}\hfill
			\subfloat[\rev{Learned: cube}]{%
				\includegraphics[width=0.475\textwidth]
				{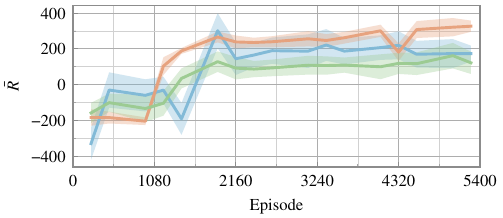}%
			}\\[-0.4em]
			\subfloat[\rev{Fixed: sphere}]{%
				\includegraphics[width=0.475\textwidth]
				{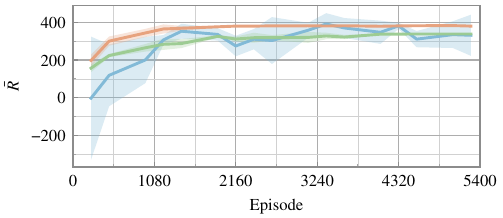}%
			}\hfill
			\subfloat[\majorrev{Fixed: nonconvex}]{%
				\includegraphics[width=0.475\textwidth]
				{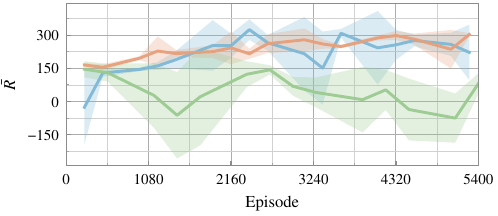}%
			}
			\\[-0.2em]
			\includegraphics[width=0.44\textwidth]
			{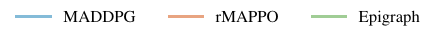}
			\caption{Training curves of normalized composite score $\bar{R}$:
				\rev{learned sphere} (a), \rev{learned cube} (b), \rev{fixed sphere} (c),
				and \majorrev{fixed nonconvex obstacle} (d).}
			\label{fig:training_curves}
		\end{figure*}

		\majorrev{Under the common task structure in Section~\ref{sec:setup}, we
			report how the human control condition affects training and real operator
			evaluation, how task and safety metrics change with geometry, and how force
			allocation changes after partner replacement.}
		
		\vspace{-0.45em}
		\subsection{\majorrev{Training Dynamics under Fixed and Learned Control}}
		
		\majorrev{Figure~\ref{fig:training_curves} shows that fixed impedance starts
			closer to late performance, reaches high scores earlier, and remains above
			joint learning through most of the budget. Under joint learning, rMAPPO
			improves most consistently, MADDPG oscillates more, and the epigraph route improves gradually; the cube increases early fluctuations without changing their
			ordering later in training.}
		
		\subsection{\majorrev{Sphere Performance by Human Control Condition}}
		
		\begin{table}[t]
			\caption{Performance in the Spherical Obstacle Scenario \rev{under Learned
					and Fixed Human Control Conditions}}
			\label{tab:perf_sphere}
			\centering
			\begin{tabular}{lcccccc}
				\hline
				Method &
				$P_{\mathrm{fin}}\uparrow$ &
				Steps$\downarrow$ &
				$\rho_{\mathrm{track}}\uparrow$ &
				$\bar{R}\uparrow$ &
				$\rho_C\downarrow$ &
				$\rho_{\mathrm{col}}\downarrow$ \\
				\hline
				\multicolumn{7}{l}{\rev{\textit{Learned condition}}} \\
				MADDPG   & 0.88          & 1200 & 0.38          & 266.19          & 0.01          & \textbf{0.00} \\
				rMAPPO   & \textbf{0.97} & 1200 & \textbf{0.61} & \textbf{353.19} & \textbf{0.00} & \textbf{0.00} \\
				Epigraph & 0.92          & 1200 & 0.03          & 119.90          & \textbf{0.00} & \textbf{0.00} \\
				\hline
				\multicolumn{7}{l}{\rev{\textit{Fixed impedance condition}}} \\
				\rev{MADDPG}   & \rev{0.97} & \rev{1200} & \rev{\textbf{0.88}} & \rev{\textbf{452.21}} & \rev{\textbf{0.00}} & \rev{\textbf{0.00}} \\
				\rev{rMAPPO}   & \rev{\textbf{0.99}} & \rev{\textbf{897}} & \rev{0.69} & \rev{383.78} & \rev{\textbf{0.00}} & \rev{\textbf{0.00}} \\
				\rev{Epigraph} & \rev{\textbf{0.99}} & \rev{1025} & \rev{0.80} & \rev{399.65} & \rev{\textbf{0.00}} & \rev{\textbf{0.00}} \\
				\hline
			\end{tabular}
			\compacttablebottom
		\end{table}
		
		\definecolor{cideal}{rgb}{0.5451,0.3529,0.1686}
		\definecolor{cmaddpg}{rgb}{0.4941,0.7216,0.8392}
		\definecolor{crmappo}{rgb}{0.9059,0.6275,0.4824}
		\definecolor{cepigraph}{rgb}{0.6078,0.7961,0.5647}
		
		\newcommand{\trajlegend}[1]{%
			\begin{tikzpicture}
				\begin{axis}[
					hide axis,
					scale only axis,
					width=0pt, height=0pt,
					xmin=0, xmax=1,
					ymin=0, ymax=1,
					legend columns=4,
					legend to name=#1,
					legend style={
						draw=none,
						fill=none,
						font=\small,
						legend cell align=left,
						legend image post style={line width=1.2pt, xscale=1.1},
						/tikz/every even column/.append style={column sep=0.9em},
					},
					]
					\addlegendimage{color=cideal, line width=1.5pt, dashed, opacity=0.55}
					\addlegendentry{Reference Path}
					\addlegendimage{color=cmaddpg, line width=2.5pt}
					\addlegendentry{MADDPG}
					\addlegendimage{color=crmappo, line width=2.5pt}
					\addlegendentry{rMAPPO}
					\addlegendimage{color=cepigraph, line width=2.5pt}
					\addlegendentry{Epigraph}
				\end{axis}
			\end{tikzpicture}%
			\pgfplotslegendfromname{#1}%
		}
		
		\majorrev{Table~\ref{tab:perf_sphere} compares the fixed and learned
			human control conditions in the spherical task.}
		\majorrev{Under fixed impedance, all methods attain high progress without
			observed exposure: MADDPG leads tracking and score, rMAPPO is shortest, and
			the epigraph route is intermediate. Joint learning separates them: rMAPPO
			provides the strongest task performance without observed violation, the
			epigraph route retains zero
			exposure but loses the most tracking and score, and MADDPG declines with
			slight Zone~C exposure. Thus partner configuration changes the operating
			point.}
		
		\majorrev{In Fig.~\ref{fig:traj_seen}(a) and (b), rMAPPO follows a compact,
			smooth bypass, MADDPG passes closer to the obstacle and re-enters more sharply,
			and the epigraph route maintains a wider bypass.}
		
		\subsection{Cross Geometry Performance
			\rev{under the Learned Condition}}
		
		We next compare the three methods under two seen geometries and one unseen
		geometry shift \rev{under the learned condition}. \rev{The transfer uses
			sphere-trained policies on a shifted cube without retraining.}
		
		% \subsubsection{Cubic-Obstacle Scenario}
		\subsubsection{Cubic Obstacle Scenario}
		
		\begin{table}[t]
			% \caption{Performance in the Cubic-Obstacle Scenario}
			\caption{\rev{Learned Condition }Performance in the Cubic Obstacle
				Scenario}
			\label{tab:perf_cube}
			\centering
			\begin{tabular}{lcccccc}
				\hline
				Method &
				$P_{\mathrm{fin}}\uparrow$ &
				Steps$\downarrow$ &
				$\rho_{\mathrm{track}}\uparrow$ &
				$\bar{R}\uparrow$ &
				$\rho_C\downarrow$ &
				$\rho_{\mathrm{col}}\downarrow$ \\
				\hline
				MADDPG   & \textbf{0.995} & \textbf{1112} & 0.368          & 208.57          & 0.012         & \textbf{0.00} \\
				rMAPPO   & 0.972          & 1200          & \textbf{0.381} & \textbf{307.50} & 0.023         & \textbf{0.00} \\
				Epigraph & 0.922          & 1200          & 0.076          & 91.22           & \textbf{0.00} & \textbf{0.00} \\
				\hline
			\end{tabular}
			\compacttablebottom
		\end{table}

		\begin{figure*}[t]
			\centering
			\subfloat[\majorrev{Sphere: front}]{%
				\includegraphics[width=0.37\textwidth]{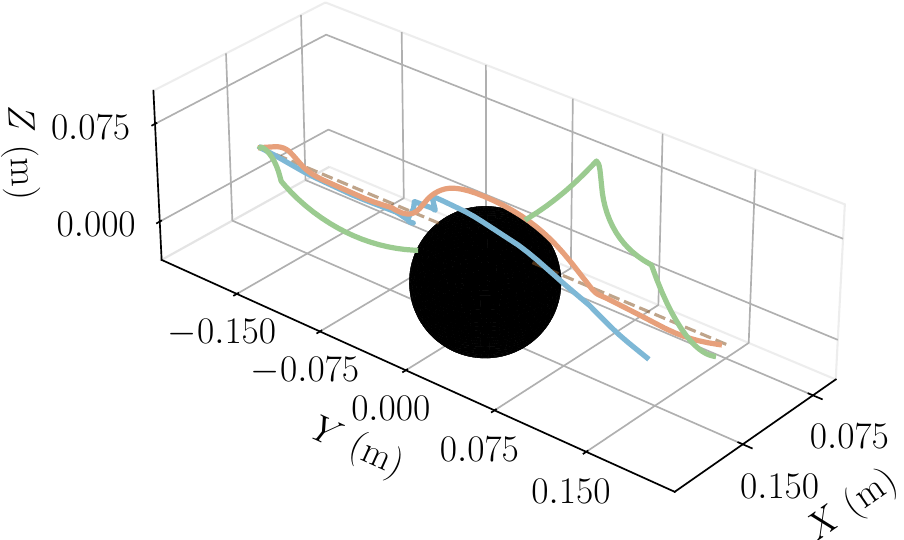}}
			\hspace{0.04\textwidth}%
			\subfloat[\majorrev{Sphere: back}]{%
				\includegraphics[width=0.37\textwidth]{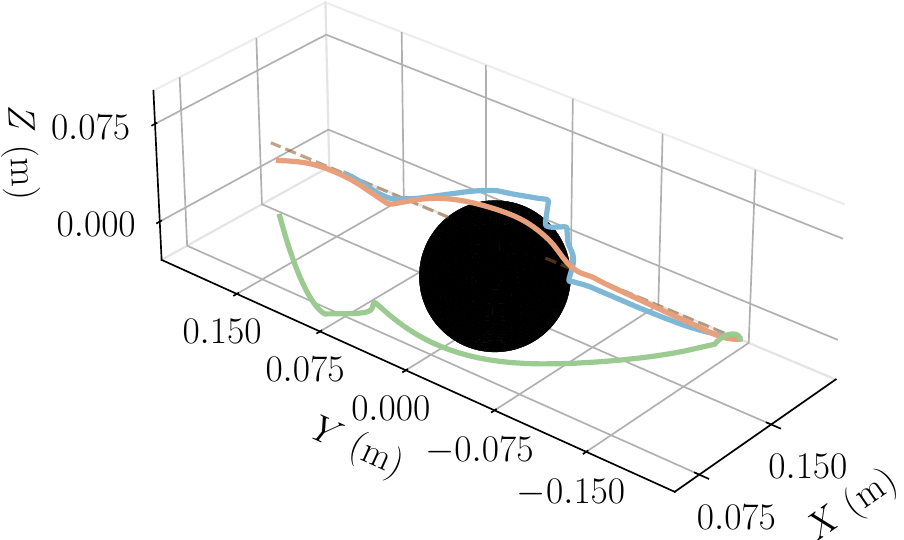}}
			\\[-0.25em]
			\subfloat[\majorrev{Cube: front}]{%
				\includegraphics[width=0.35\textwidth]{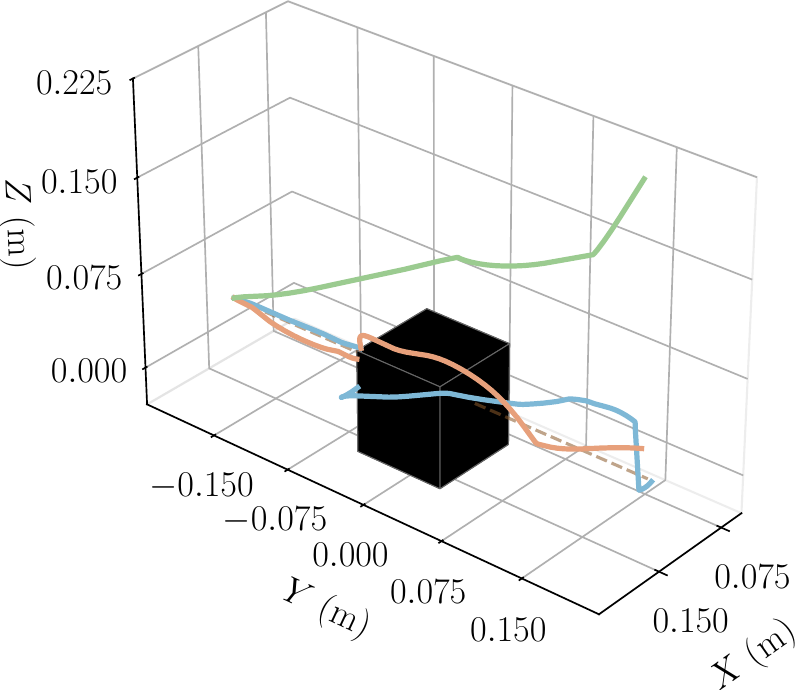}}
			\hspace{0.04\textwidth}%
			\subfloat[\majorrev{Cube: back}]{%
				\includegraphics[width=0.35\textwidth]{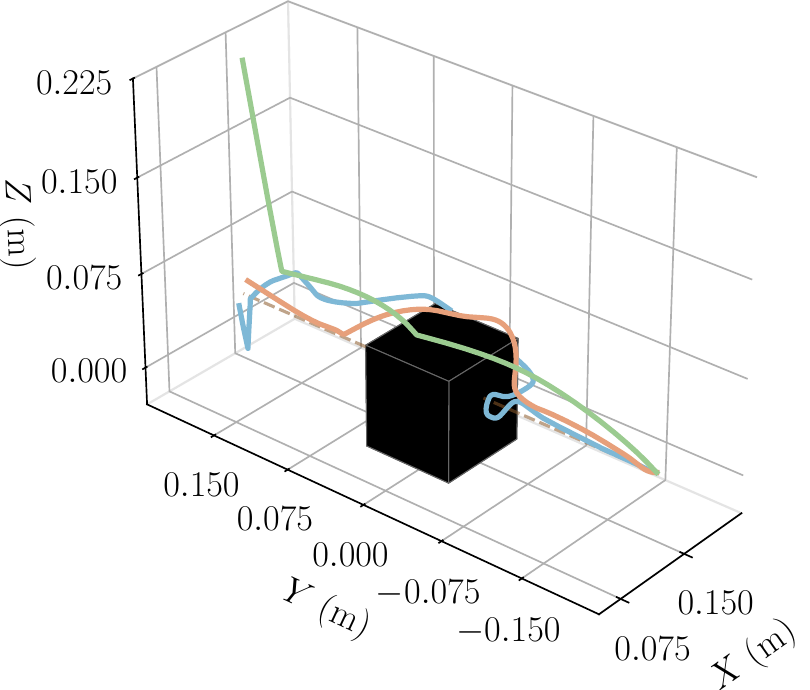}}
			\vspace{0.10em}
			\trajlegend{trajlegend_cube}
			\caption{\majorrev{Learned condition trajectories for the spherical and
					cubic obstacles.}}
			\label{fig:traj_seen}
		\end{figure*}
		
		Table~\ref{tab:perf_cube} shows a similar ordering in the cubic-obstacle
		scenario. rMAPPO again achieves the highest observed composite score and tracking
		ratio, \rev{although both decrease from sphere to cube and its Zone~C
			exposure becomes nonzero}. MADDPG attains the highest final progress and
		the fewest steps, but with a lower composite score and tracking quality. The epigraph route again
		shows no observed Zone~C or collision exposure, but has the lowest
		composite score. \rev{Across the two seen geometries, rMAPPO preserves its
			task performance advantage, the epigraph route its zero exposure safety behavior, and MADDPG shows marked geometry-dependent variation.}
		
		\majorrev{The corresponding trajectories in Fig.~\ref{fig:traj_seen}(c) and
			(d) follow the same pattern: rMAPPO forms a smooth bypass, MADDPG travels
			closer to the cube corners, and the epigraph route keeps a wider separation from the
			obstacle.}

		% \subsubsection{Geometric Generalization Scenario}
		\subsubsection{Geometric Generalization
			\rev{under the Learned Condition}}
		
		\begin{table}[t]
			\caption{Generalization Performance \rev{under the Learned Condition}:
				Trained on Sphere, Tested on Shifted Cube}
			\label{tab:perf_gen}
			\centering
			\begin{tabular}{lcccccc}
				\hline
				Method &
				$P_{\mathrm{fin}}\uparrow$ &
				Steps$\downarrow$ &
				$\rho_{\mathrm{track}}\uparrow$ &
				$\bar{R}\uparrow$ &
				$\rho_C\downarrow$ &
				$\rho_{\mathrm{col}}\downarrow$ \\
				\hline
				MADDPG   & 0.596          & 1200 & \textbf{0.332} & 123.99          & \textbf{0.00} & \textbf{0.00} \\
				rMAPPO   & \textbf{0.956} & 1200 & 0.304          & \textbf{293.51} & 0.051         & 0.011 \\
				Epigraph & 0.934          & 1200 & 0.028          & 127.19          & \textbf{0.00} & \textbf{0.00} \\
				\hline
			\end{tabular}
			\compacttablebottom
		\end{table}
		
		As Table~\ref{tab:perf_gen} illustrates, rMAPPO still achieves the highest final progress and composite score after direct transfer, but it is also the only method with nonzero collision and Zone~C exposure. MADDPG shows the weakest transfer in final progress. The epigraph route, by contrast, shows no observed collision or Zone~C exposure, but has a lower tracking ratio. \rev{The transferred policies therefore differ primarily in retained task performance and geometric safety.}
		
		\majorrev{As Fig.~\ref{fig:traj_transfer_nonconvex}(c) and (d) show, after the
			cube is shifted forward, rMAPPO still adjusts its path in advance but part of
			the trajectory enters the obstacle. MADDPG remains in front of the obstacle
			for a prolonged period, whereas the epigraph route selects a wider bypass and remains
			outside the obstacle.}
		
		\subsection{\majorrev{Rigid Nonconvex Geometry}}
		
		\majorrev{Figure~\ref{fig:training_curves}(d) compares nonconvex optimization
			shapes: rMAPPO improves comparatively steadily, MADDPG is noisier, and the
			epigraph route has high variability across seeds.}

		\begin{table}[t]
			\caption{Performance on the Rigid Nonconvex Geometry}
			\label{tab:perf_nonconvex}
			\centering
			\setlength{\tabcolsep}{3pt}
			\begin{tabular}{lccc}
				\hline
				\noalign{\vskip 2pt}
				\makecell[l]{\textit{Task behavior}\\Method} &
				\makecell[c]{$P_{\mathrm{fin}}\uparrow$} &
				\makecell[c]{$\rho_{\mathrm{track}}\uparrow$} &
				\makecell[c]{Traj. RMSE (mm)$\downarrow$} \\
				\noalign{\vskip 2pt}
				\hline
				MADDPG   & $0.576\pm0.353$ & $\mathbf{0.878\pm0.142}$ & $11.26\pm8.64$ \\
				rMAPPO   & $\mathbf{0.996\pm0.003}$ & $0.796\pm0.140$ & $\mathbf{7.87\pm0.64}$ \\
				Epigraph & $0.995\pm0.001$ & $0.359\pm0.222$ & $25.77\pm4.44$ \\
				\hline
			\end{tabular}
			
			% \vspace{0.35em}
			\vspace{0.7em}
			\begin{tabular}{lcc}
				\hline
				\noalign{\vskip 2pt}
				\makecell[l]{\textit{Geometric safety}\\Method} &
				\makecell[c]{$\rho_C\downarrow$} &
				\makecell[c]{$\rho_{\mathrm{col}}\downarrow$} \\
				\noalign{\vskip 2pt}
				\hline
				MADDPG   & $0.0056\pm0.0096$ & $\mathbf{0.0000\pm0.0000}$ \\
				rMAPPO   & $0.0434\pm0.0014$ & $0.0175\pm0.0008$ \\
				Epigraph & $\mathbf{0.0025\pm0.0043}$ & $\mathbf{0.0000\pm0.0000}$ \\
				\hline
			\end{tabular}
			\compacttablebottom
		\end{table}
		
		\majorrev{Table~\ref{tab:perf_nonconvex} separates later task and safety behavior. All runs reach the 1199-step horizon without satisfying the strict $10\,\mathrm{mm}$ success criterion. rMAPPO and the epigraph route finish only $10.17$ to $10.93\,\mathrm{mm}$ and $10.24$ to $11.28\,\mathrm{mm}$ from the goal, respectively. rMAPPO tracks more closely but has the greatest Zone~C and collision exposure; the epigraph route retains comparable progress with the lowest Zone~C exposure and no collision. MADDPG records a high raw tracking ratio, but two seeds stop $248$ to $267\,$mm from the goal, so this value reflects remaining near the nominal line rather than completing it. Across the three seeds, the epigraph route therefore provides collision-free near-completion with the lowest aggregate Zone~C exposure, albeit with wider trajectory deviations. The comparison across three seeds remains descriptive.}
		
		\begin{figure*}[!t]
			\centering
			\subfloat[\majorrev{Nonconvex: front}]{%
				\includegraphics[width=0.39\textwidth]{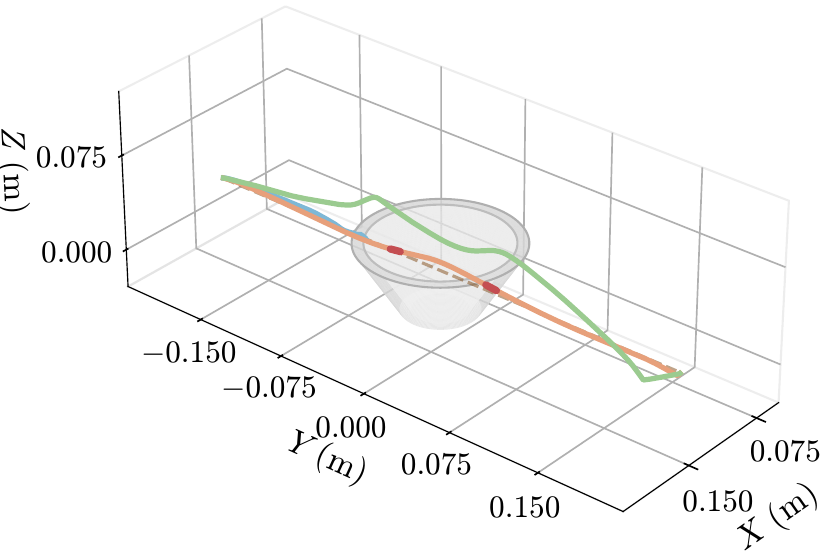}}
			\hspace{0.015\textwidth}%
			\subfloat[\majorrev{Nonconvex: back}]{%
				\includegraphics[width=0.39\textwidth]{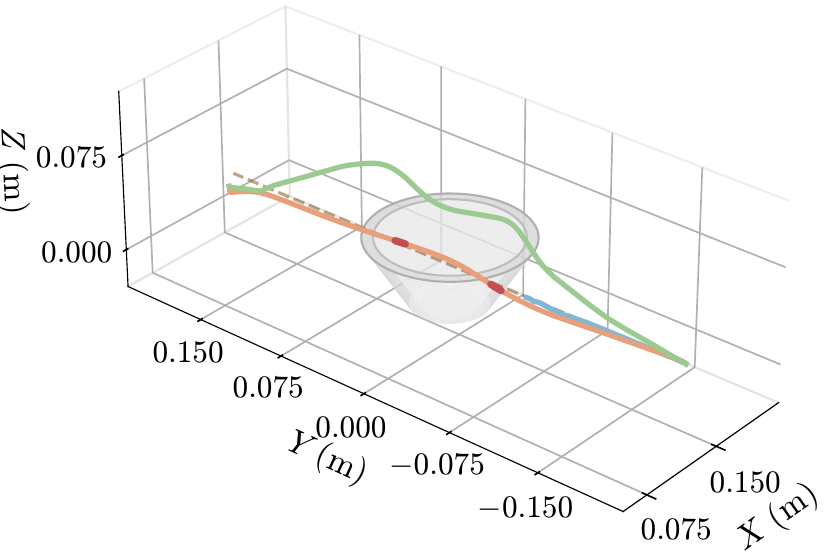}}
			\\[-0.25em]
			\subfloat[\majorrev{Transfer: front}]{%
				\includegraphics[width=0.37\textwidth]{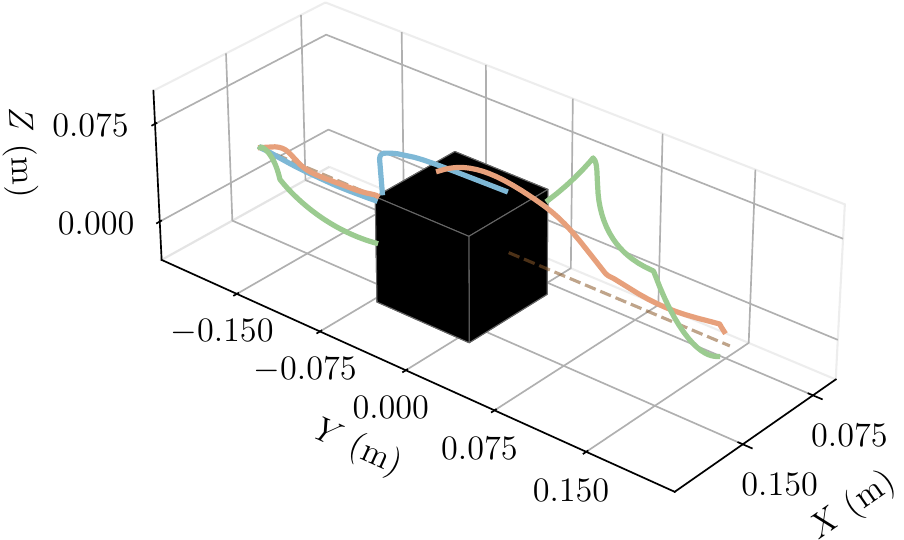}}
			\hspace{0.04\textwidth}%
			\subfloat[\majorrev{Transfer: back}]{%
				\includegraphics[width=0.37\textwidth]{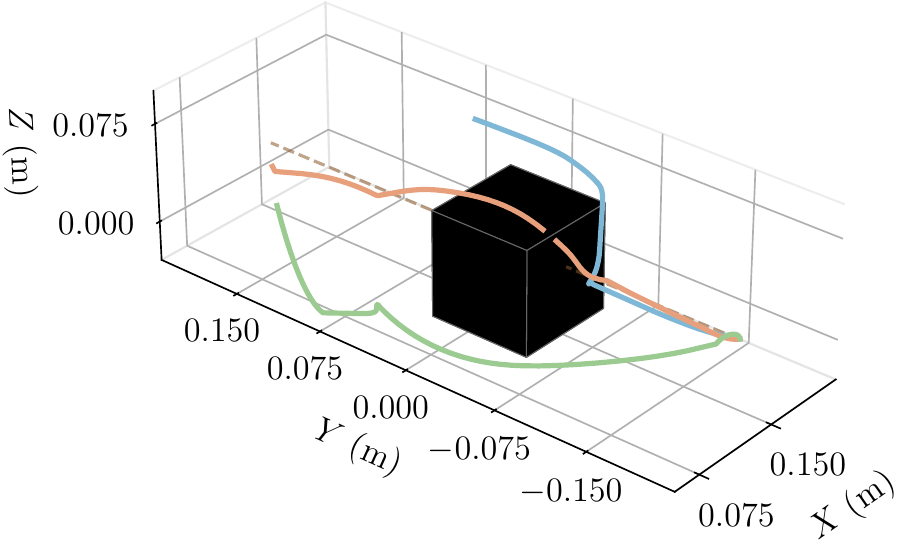}}
			\vspace{0.10em}
			\trajlegend{trajlegend_gen}
			\caption{\majorrev{Rigid nonconvex and sphere to cube transfer trajectories;
					red segments indicate collision.}}
			\label{fig:traj_transfer_nonconvex}
		\end{figure*}
		
		\majorrev{In Fig.~\ref{fig:traj_transfer_nonconvex}(a) and (b), MADDPG stops
			before completing the route, rMAPPO remains near the reference but intersects
			the obstacle, and the epigraph route takes a wider vertical bypass without
			collision.}
		
		\subsection{\majorrev{Simulation Force Coordination}}
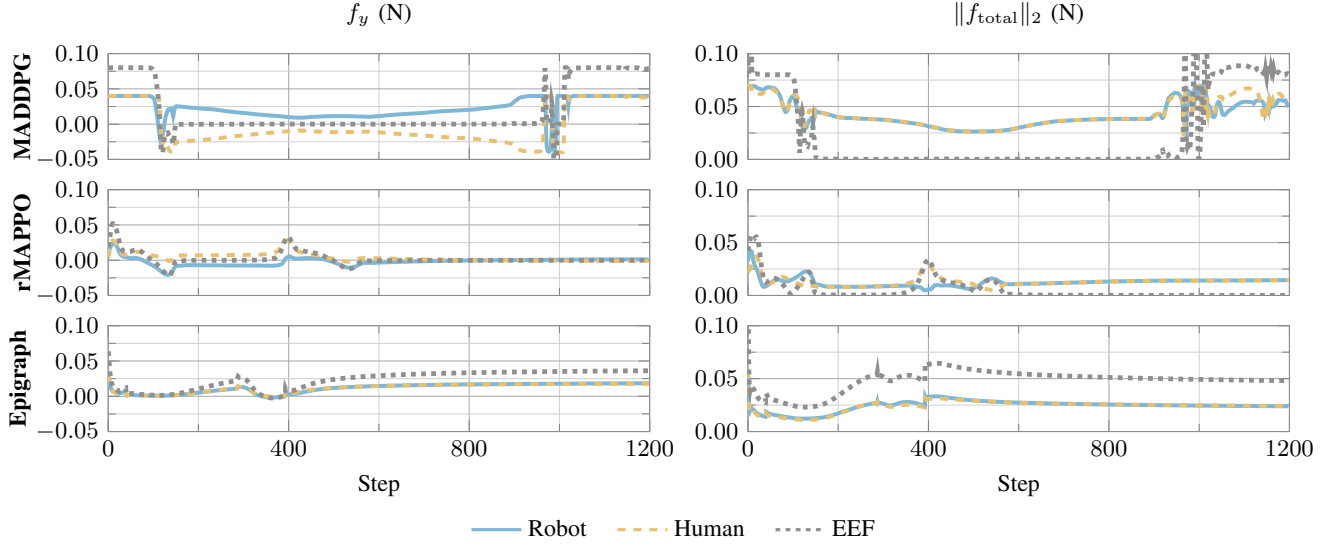
\begin{figure*}[!t]
			\centering
			
			\definecolor{mycolorhuman}{rgb}{0.9098,0.7569,0.4392}
			\definecolor{mycolorrobot}{rgb}{0.4941,0.7216,0.8392}
			\definecolor{mycolorcombine}{rgb}{0.5608,0.5608,0.5608}
			
			\pgfplotsset{
				forcegroup/.style={
					scale only axis,
					width=0.395\textwidth,
					height=1.4cm,
					title style={font=\small, yshift=0.2ex},
					xlabel style={font=\small},
					ylabel style={font=\small},
					tick label style={font=\small},
					grid=both,
					grid style={line width=0.25pt, draw=gridlight},
					major grid style={line width=0.4pt, draw=gridmajor},
					minor tick num=1,
					xmin=0, xmax=1200,
					xtick={0,400,800,1200},
					xticklabel style={
						/pgf/number format/1000 sep={},
						font=\small,
					},
					yticklabel style={
						/pgf/number format/fixed,
						/pgf/number format/fixed zerofill,
						/pgf/number format/precision=2,
						font=\small,
					},
					scaled x ticks=false,
					scaled y ticks=false,
					axis line style={axisgray},
					clip=true,
				}
			}
			
			\makebox[\textwidth][c]{%
				\begin{tikzpicture}
					\begin{groupplot}[
						group style={
							group size=2 by 3,
							horizontal sep=1.3cm,
							vertical sep=0.40cm,
						},
						forcegroup,
						]
						
						\nextgroupplot[
						title={$f_y$ (N)},
						ymin=-0.05, ymax=0.10,
						ytick={-0.05,0,0.05,0.10},
						xticklabels=\empty,
						xlabel={},
						legend to name=forcelegend,
						legend columns=3,
						legend style={
							draw=none,
							fill=none,
							font=\small,
							legend cell align=left,
							legend image post style={line width=1.2pt, xscale=1.1},
							/tikz/every even column/.append style={column sep=0.9em},
						},
						]
						\addplot[color=mycolorrobot, line width=1.5pt, mark=none]
						table[x=step, y=robot_fy, col sep=comma]
						{data_tikz/forces_combine_maddpg.csv};
						\addlegendentry{Robot}
						
						\addplot[color=mycolorhuman, line width=1.5pt, mark=none, dashed]
						table[x=step, y=human_fy, col sep=comma]
						{data_tikz/forces_combine_maddpg.csv};
						\addlegendentry{Human}
						
						\addplot[color=mycolorcombine, line width=1.80pt, mark=none, dotted]
						table[x=step, y=robot_human_fy_combine, col sep=comma]
						{data_tikz/forces_combine_maddpg.csv};
						\addlegendentry{EEF}
						
						\nextgroupplot[
						title={$\lVert f_{\mathrm{total}}\rVert_2$ (N)},
						ymin=0, ymax=0.10,
						ytick={0,0.05,0.10},
						xticklabels=\empty,
						xlabel={},
						]
						\addplot[color=mycolorrobot, line width=1.5pt, mark=none]
						table[x=step, y=robot_f_norm, col sep=comma]
						{data_tikz/forces_combine_maddpg.csv};
						\addplot[color=mycolorhuman, line width=1.5pt, mark=none, dashed]
						table[x=step, y=human_f_norm, col sep=comma]
						{data_tikz/forces_combine_maddpg.csv};
						\addplot[color=mycolorcombine, line width=1.80pt, mark=none, dotted]
						table[x=step, y=robot_human_f_norm_combine, col sep=comma]
						{data_tikz/forces_combine_maddpg.csv};
						
						\nextgroupplot[
						ymin=-0.05, ymax=0.10,
						ytick={-0.05,0,0.05,0.10},
						xticklabels=\empty,
						xlabel={},
						]
						\addplot[color=mycolorrobot, line width=1.5pt, mark=none]
						table[x=step, y=robot_fy, col sep=comma]
						{data_tikz/forces_combine_rmappo.csv};
						\addplot[color=mycolorhuman, line width=1.5pt, mark=none, dashed]
						table[x=step, y=human_fy, col sep=comma]
						{data_tikz/forces_combine_rmappo.csv};
						\addplot[color=mycolorcombine, line width=1.80pt, mark=none, dotted]
						table[x=step, y=robot_human_fy_combine, col sep=comma]
						{data_tikz/forces_combine_rmappo.csv};
						
						\nextgroupplot[
						ymin=0, ymax=0.10,
						ytick={0,0.05,0.10},
						xticklabels=\empty,
						xlabel={},
						]
						\addplot[color=mycolorrobot, line width=1.5pt, mark=none]
						table[x=step, y=robot_f_norm, col sep=comma]
						{data_tikz/forces_combine_rmappo.csv};
						\addplot[color=mycolorhuman, line width=1.5pt, mark=none, dashed]
						table[x=step, y=human_f_norm, col sep=comma]
						{data_tikz/forces_combine_rmappo.csv};
						\addplot[color=mycolorcombine, line width=1.80pt, mark=none, dotted]
						table[x=step, y=robot_human_f_norm_combine, col sep=comma]
						{data_tikz/forces_combine_rmappo.csv};
						
						\nextgroupplot[
						xlabel={Step},
						ymin=-0.05, ymax=0.10,
						ytick={-0.05,0,0.05,0.10},
						]
						\addplot[color=mycolorrobot, line width=1.5pt, mark=none]
						table[x=step, y=robot_fy, col sep=comma]
						{data_tikz/forces_combine_epigraph.csv};
						\addplot[color=mycolorhuman, line width=1.5pt, mark=none, dashed]
						table[x=step, y=human_fy, col sep=comma]
						{data_tikz/forces_combine_epigraph.csv};
						\addplot[color=mycolorcombine, line width=1.80pt, mark=none, dotted]
						table[x=step, y=robot_human_fy_combine, col sep=comma]
						{data_tikz/forces_combine_epigraph.csv};
						
						\nextgroupplot[
						xlabel={Step},
						ymin=0, ymax=0.10,
						ytick={0,0.05,0.10},
						]
						\addplot[color=mycolorrobot, line width=1.5pt, mark=none]
						table[x=step, y=robot_f_norm, col sep=comma]
						{data_tikz/forces_combine_epigraph.csv};
						\addplot[color=mycolorhuman, line width=1.5pt, mark=none, dashed]
						table[x=step, y=human_f_norm, col sep=comma]
						{data_tikz/forces_combine_epigraph.csv};
						\addplot[color=mycolorcombine, line width=1.80pt, mark=none, dotted]
						table[x=step, y=robot_human_f_norm_combine, col sep=comma]
						{data_tikz/forces_combine_epigraph.csv};
						
					\end{groupplot}
					
					\node[overlay, rotate=90, anchor=south, font=\small]
					at ($(group c1r1.west)+(-0.90cm,0)$) {\textbf{MADDPG}};
					\node[overlay, rotate=90, anchor=south, font=\small]
					at ($(group c1r2.west)+(-0.90cm,0)$) {\textbf{rMAPPO}};
					\node[overlay, rotate=90, anchor=south, font=\small]
					at ($(group c1r3.west)+(-0.90cm,0)$) {\textbf{Epigraph}};
				\end{tikzpicture}%
			}
			
			\vspace{0.15em}
			\pgfplotslegendfromname{forcelegend}
			\caption{Force comparison across MADDPG, rMAPPO, and Epigraph over one
				representative \rev{spherical obstacle} episode \rev{under the learned
					condition}. The left column shows the $y$-axis force components of robot,
				human, and their combined force. The right column shows the robot force
				magnitude, human force magnitude, and the magnitude of the combined
				human-robot force.}
			\label{fig:force_methods_3x2}
		\end{figure*}

		\majorrev{Figure~\ref{fig:force_methods_3x2} shows representative learned
			sphere episodes. MADDPG and rMAPPO often produce opposite Cartesian
			components and nonuniform magnitudes, consistent with different roles under
			their agent-specific critics. The epigraph forces are more synchronized; its
			actors share a performance critic while retaining agent-specific safety
			critics. These profiles show association, not causality or universal
			superiority of either allocation.}
		
		\subsection{Real Operator Evaluation}
		
		\begin{table*}[!t]
			\caption{Human in the Loop Task and Safety Performance}
			\label{tab:hil_task}
			\centering
			\setlength{\tabcolsep}{2.2pt}
			\begin{tabular}{llcccccc}
				\hline
				\noalign{\vskip 3pt}
				\makecell[l]{Method} & \makecell[c]{Training\\condition} &
				\makecell[c]{Success$\uparrow$} & \makecell[c]{Steps$\downarrow$} &
				\makecell[c]{$\rho_{\mathrm{track}}\uparrow$} &
				\makecell[c]{$\rho_C\downarrow$} &
				\makecell[c]{$\rho_{\mathrm{col}}\downarrow$} &
				\makecell[c]{Device-to-EEF\\tracking RMSE (mm)$\downarrow$} \\
				\noalign{\vskip 3pt}
				\hline
				MADDPG   & Fixed     & $\mathbf{1.00\pm0.00}$ & $551\pm45$  & $0.226\pm0.121$ & $0.0369\pm0.0795$ & $0.0194\pm0.0428$ & $29.9\pm10.7$ \\
				MADDPG   & Learned & $0.90\pm0.32$ & $787\pm185$ & $0.101\pm0.074$ & $0.0110\pm0.0208$ & $\mathbf{0.0000\pm0.0000}$ & $79.6\pm17.1$ \\
				rMAPPO   & Fixed     & $\mathbf{1.00\pm0.00}$ & $625\pm90$  & $\mathbf{0.238\pm0.157}$ & $0.0414\pm0.0698$ & $0.0206\pm0.0486$ & $\mathbf{17.4\pm5.0}$ \\
				rMAPPO   & Learned & $0.70\pm0.48$ & $899\pm216$ & $\mathbf{0.311\pm0.083}$ & $0.0366\pm0.0693$ & $0.0038\pm0.0112$ & $\mathbf{29.8\pm4.1}$ \\
				Epigraph & Fixed     & $\mathbf{1.00\pm0.00}$ & $\mathbf{543\pm84}$  & $0.197\pm0.043$ & $\mathbf{0.0018\pm0.0056}$ & $\mathbf{0.0000\pm0.0000}$ & $57.0\pm32.0$ \\
				Epigraph & Learned & $\mathbf{1.00\pm0.00}$ & $\mathbf{668\pm172}$ & $0.207\pm0.057$ & $\mathbf{0.0097\pm0.0221}$ & $\mathbf{0.0000\pm0.0000}$ & $52.1\pm15.8$ \\
				\hline
			\end{tabular}
			\compacttablebottom
		\end{table*}

		\majorrev{Table~\ref{tab:hil_task} shows $10/10$ success and fewer steps for all policies trained with fixed impedance. With real operator input, the MADDPG and rMAPPO policies trained with a learned partner fall to $0.90$ and $0.70$ success, whereas both epigraph policies retain full success, zero collision, and the lowest Zone~C exposure. Fixed rMAPPO has the lowest device-to-EEF tracking RMSE; learned rMAPPO has the highest tracking ratio despite its lower completion reliability.}
		
		\begin{table*}[!t]
			\caption{Human in the Loop Force Allocation}
			\label{tab:hil_force}
			\centering
			\setlength{\tabcolsep}{4.0pt}
			\begin{tabular}{llcccc}
				\hline
				\noalign{\vskip 3pt}
				\makecell[l]{Method} & \makecell[c]{Training\\condition} &
				\makecell[c]{Human share\\$\eta_H$} &
				\makecell[c]{Mean cosine\\$\bar\kappa$} &
				\makecell[c]{Opposing ratio\\$\rho_{\mathrm{opp}}$} &
				\makecell[c]{Cancellation\\$\chi$} \\
				\noalign{\vskip 3pt}
				\hline
				MADDPG   & Fixed     & $0.398\pm0.030$ & $-0.055\pm0.103$ & $0.592\pm0.065$ & $0.305\pm0.034$ \\
				MADDPG   & Learned & $0.483\pm0.008$ & $-0.740\pm0.126$ & $\mathbf{0.898\pm0.082}$ & $0.743\pm0.084$ \\
				rMAPPO   & Fixed     & $0.525\pm0.034$ & $-0.317\pm0.161$ & $0.688\pm0.112$ & $0.421\pm0.092$ \\
				rMAPPO   & Learned & $0.509\pm0.007$ & $-0.951\pm0.022$ & $0.997\pm0.003$ & $0.858\pm0.029$ \\
				Epigraph & Fixed     & $0.523\pm0.026$ & $\mathbf{\phantom{-}0.314\pm0.182}$ & $\mathbf{0.260\pm0.121}$ & $\mathbf{0.178\pm0.056}$ \\
				Epigraph & Learned & $0.634\pm0.040$ & $\mathbf{-0.659\pm0.067}$ & $0.939\pm0.038$ & $\mathbf{0.539\pm0.058}$ \\
				\hline
			\end{tabular}
			\compacttablebottom
		\end{table*}

		\majorrev{Table~\ref{tab:hil_force} shows that fixed impedance with the
			epigraph route is the only condition with a positive mean cosine and has the
			least opposition and cancellation. For every method, the policy trained with
			a learned partner produces more opposing and cancelling forces with real
			operator input; rMAPPO reaches $0.997$ and $0.858$, respectively. Human share
			ranges from $0.398$ to $0.634$.}
		
		\section{\majorrev{Discussions and Limitations}}
		\label{sec:discussion}
		
		\majorrev{The fixed and learned conditions are motivated by the optimal control theory of the human operator, but the study does not evaluate the fidelity or the physiological descriptions of human behavior. 
			Fixed impedance provides a task reference, while the learned policy is an adaptive computational partner. Its lower efficiency in several settings shows that joint adaptation is not automatically beneficial. The real operator study therefore evaluates robot policy compatibility. Although the fixed versus learned efficiency trend persists, method rankings and force allocation change, showing that coordination learned with a computational partner need not transfer unchanged to real operators.}
		
		\majorrev{The results reveal geometry- and metric-dependent differences among the methods. rMAPPO often achieves stronger progress or tracking, whereas the epigraph route more consistently limits observed exposure, although its wider avoidance can reduce efficiency. No method consistently dominates in the nonconvex task under the available training budgets. A potential extension is to combine learned constraint-aware policies with CBF or QP safety filters, subject to suitable interaction models and feasible runtime corrections.}
		
		\majorrev{The nonconvex study uses a rigid object, three seeds, unequal available training horizons, and one evaluation episode per seed; it is not a comparison at matched convergence and does not represent deformable tissue. The broader simulation comparison also uses limited geometry and randomization, and its exposure rates are empirical rather than formal safety guarantees. Controller form and the number of trained actors differ across complete configurations, so the data do not isolate one architectural factor.}
		
		\majorrev{The real operator evidence is descriptive: it covers two operators,
			one spherical task, five episodes per condition, and one robot checkpoint per
			condition. The same operators contributed the motion trials used to calibrate
			the device coupling, so this is not an independent validation cohort. The
			results support feasibility and compatibility analysis, not population or
			clinical claims. The Phantom provides no direct force measurement, so the
			human command is reconstructed through the calibrated coupling controller.}

		\section{\majorrev{Conclusion and Future Work}} 
		\majorrev{This work established a simulation and real-operator framework for safety-aware MARL in haptic surgical shared control. MADDPG, rMAPPO, and an epigraph-constrained route were compared with fixed and learned training partners on the spherical task. Learned policies were also tested on cubic and transferred geometries, and fixed-impedance policies on nonconvex geometry. All methods used common observations and geometric metrics but route-specific objectives.} 
		\majorrev{In simulation, rMAPPO generally achieved the highest composite scores and tracking ratios, while MADDPG was more sensitive to geometry and training condition. The epigraph route most consistently limited observed exposure, although wider avoidance sometimes reduced efficiency. Where directly comparable, fixed-impedance training generally produced higher scores and shorter execution than joint learning.} 
		\majorrev{In the descriptive two-operator study, all fixed-impedance-trained policies completed every recorded trial. Learned-partner training reduced MADDPG and rMAPPO reliability and increased opposition between robot and reconstructed operator-side commands. Both epigraph policies completed every trial, and the fixed condition showed the most aligned commands. Thus, coordination learned with a computational partner may not persist with real operator input.} 
		
		\majorrev{Future work will evaluate larger independent cohorts, randomized conditions, direct force sensing, broader geometries, deformable interaction, full robot dynamics, latency sensitivity, and comparisons with CBF/QP safety filters.}



\begin{thebibliography}{99}
			\providecommand{\url}[1]{#1}
			\csname url@samestyle\endcsname
			\providecommand{\newblock}{\relax}
			\providecommand{\bibinfo}[2]{#2}
			\providecommand{\BIBentrySTDinterwordspacing}{\spaceskip=0pt\relax}
			\providecommand{\BIBentryALTinterwordstretchfactor}{4}
			\providecommand{\BIBentryALTinterwordspacing}{\spaceskip=\fontdimen2\font plus
				\BIBentryALTinterwordstretchfactor\fontdimen3\font minus
				\fontdimen4\font\relax}
			\providecommand{\BIBforeignlanguage}[2]{{%
					\expandafter\ifx\csname l@#1\endcsname\relax
					\typeout{** WARNING: IEEEtran.bst: No hyphenation pattern has been}%
					\typeout{** loaded for the language `#1'. Using the pattern for}%
					\typeout{** the default language instead.}%
					\else
					\language=\csname l@#1\endcsname
					\fi
					#2}}
			\providecommand{\BIBdecl}{\relax}
			\BIBdecl
			
			\bibitem{Li2023}
			G.~Li, Q.~Li, C.~Yang, Y.~Su, Z.~Yuan, and X.~Wu, ``The classification and new
			trends of shared control strategies in telerobotic systems: A survey,''
			\emph{IEEE Transactions on Haptics}, vol.~16, no.~2, pp. 118--133, 2023.
			
			\bibitem{Chen2024}
			L.~Chen, Z.~J. Hu, Y.~Huang, E.~Burdet, and F.~Rodriguez~y Baena, ``Human robot
			shared control in surgery: A performance assessment,'' in \emph{2024 IEEE
				International Conference on Robotics and Automation (ICRA)}, 2024, pp.
			15\,165--15\,171.
			
			\bibitem{Ebrahimi2021TBME}
			A.~Ebrahimi, M.~G. Urias, N.~A. Patel, R.~H. Taylor, P.~L. Gehlbach, and
			I.~Iordachita, ``Adaptive control improves sclera force safety in
			robot-assisted eye surgery: A clinical study,'' \emph{IEEE Transactions on
				Biomedical Engineering}, vol.~68, no.~11, pp. 3356--3365, 2021.
			
			\bibitem{Patel2022}
			R.~V. Patel, S.~F. Atashzar, and M.~Tavakoli, ``Haptic feedback and force-based
			teleoperation in surgical robotics,'' \emph{Proceedings of the IEEE}, vol.
			110, no.~7, pp. 1012--1027, 2022.
			
			\bibitem{Brunke2021}
			L.~Brunke, M.~Greeff, A.~W. Hall, Z.~Yuan, S.~Zhou, J.~Panerati, and A.~P.
			Schoellig, ``Safe learning in robotics: From learning-based control to safe
			reinforcement learning,'' \emph{Annual Review of Control, Robotics, and
				Autonomous Systems}, vol.~5, no.~1, pp. 411--444, 2022.
			
			\bibitem{Yang2017}
			G.-Z. Yang, J.~Cambias, K.~Cleary, E.~Daimler, J.~Drake, P.~E. Dupont, N.~Hata,
			P.~Kazanzides, S.~Martel, R.~V. Patel, V.~J. Santos, and R.~H. Taylor,
			``Medical robotics---regulatory, ethical, and legal considerations for
			increasing levels of autonomy,'' \emph{Science Robotics}, vol.~2, no.~4, p.
			eaam8638, 2017.
			
			\bibitem{Lee2024}
			A.~Lee, T.~Baker, J.~Bederson, and B.~Rapoport, ``Levels of autonomy in
			{FDA}-cleared surgical robots: A systematic review,'' \emph{npj Digital
				Medicine}, vol.~7, no.~1, 2024.
			
			\bibitem{Suresh2024}
			P.~S. Suresh, D.~Romeres, P.~Doshi, and S.~Jain, ``Open human-robot
			collaboration systems ({OHRCS}): A research perspective,'' in \emph{2024 IEEE
				6th International Conference on Cognitive Machine Intelligence (CogMI)},
			2024, pp. 103--110.
			
			\bibitem{Tian2023TMRB}
			H.~Tian, Z.~Han, Y.~Wang, X.~Zhu, W.~Zhang, Z.~Wang, C.~Li, and X.~Duan,
			``Virtual-fixture-based osteotomy shared control: A framework for human-robot
			shared surgical osteotomy manipulation,'' \emph{IEEE Transactions on Medical
				Robotics and Bionics}, vol.~5, no.~4, pp. 945--955, 2023.
			
			\bibitem{Varga2025Sensors}
			B.~Varga and M.~Poncelet, ``A shared control approach to robot-assisted
			cataract surgery training for novice surgeons,'' \emph{Sensors}, vol.~25,
			no.~16, p. 5165, 2025.
			
			\bibitem{Li2019}
			\BIBentryALTinterwordspacing
			Y.~Li, K.~P. Tee, R.~Yan, and S.~S. Ge, ``Reinforcement learning for
			human-robot shared control,'' \emph{Assembly Automation}, vol.~40, no.~1, pp.
			105--117, 2020. [Online]. Available:
			\url{https://doi.org/10.1108/AA-10-2018-0153}
			\BIBentrySTDinterwordspacing
			
			\bibitem{Music2020}
			S.~Musi{\'c} and S.~Hirche, ``Haptic shared control for human-robot
			collaboration: A game-theoretical approach,'' \emph{IFAC-PapersOnLine},
			vol.~53, no.~2, pp. 10\,216--10\,222, 2020.
			
			\bibitem{Qin2025}
			W.~Qin, H.~Yi, Z.~Fan, and J.~Zhao, ``Haptic shared control framework with
			interaction force constraint based on control barrier function for
			teleoperation,'' \emph{Sensors}, vol.~25, no.~2, p. 405, 2025.
			
			\bibitem{Su2019RAL}
			H.~Su, C.~Yang, G.~Ferrigno, and E.~De~Momi, ``Improved human--robot
			collaborative control of redundant robot for teleoperated minimally invasive
			surgery,'' \emph{IEEE Robotics and Automation Letters}, vol.~4, no.~2, pp.
			1447--1453, 2019.
			
			\bibitem{Enayati2016}
			N.~Enayati, E.~De~Momi, and G.~Ferrigno, ``Haptics in robot-assisted surgery:
			Challenges and benefits,'' \emph{IEEE Reviews in Biomedical Engineering},
			vol.~9, pp. 49--65, 2016.
			
			\bibitem{Bergholz2023}
			M.~Bergholz, M.~Ferle, and B.~M. Weber, ``The benefits of haptic feedback in
			robot assisted surgery and their moderators: A meta-analysis,''
			\emph{Scientific Reports}, vol.~13, p. 19215, 2023.
			
			\bibitem{Scheikl2021}
			P.~M. Scheikl, B.~Gyenes, T.~Davitashvili, R.~Younis, A.~Schulze, B.~P.
			Muller-Stich, G.~Neumann, M.~Wagner, and F.~Mathis-Ullrich, ``Cooperative
			assistance in robotic surgery through multi-agent reinforcement learning,''
			in \emph{2021 IEEE/RSJ International Conference on Intelligent Robots and
				Systems (IROS)}.\hskip 1em plus 0.5em minus 0.4em\relax IEEE, Sep. 2021, pp.
			1859--1864.
			
			\bibitem{varga_toward_2024}
			B.~Varga, ``Toward {Adaptive} {Cooperation}: {Model}-{Based} {Shared} {Control}
			{Using} {LQ}-{Differential} {Games},'' \emph{Acta Polytechnica Hungarica},
			vol.~21, no.~10, pp. 439--456, 2024.
			
			\bibitem{Hu2023}
			Z.~J. Hu, Z.~Wang, Y.~Huang, A.~Sena, F.~Rodriguez~y Baena, and E.~Burdet,
			``Towards human-robot collaborative surgery: Trajectory and strategy learning
			in bimanual peg transfer,'' \emph{IEEE Robotics and Automation Letters},
			vol.~8, no.~8, pp. 4553--4560, 2023.
			
			\bibitem{Su2021TASE}
			H.~Su, A.~Mariani, S.~E. Ovur, A.~Menciassi, G.~Ferrigno, and E.~De~Momi,
			``Toward teaching by demonstration for robot-assisted minimally invasive
			surgery,'' \emph{IEEE Transactions on Automation Science and Engineering},
			vol.~18, no.~2, pp. 484--494, 2021.
			
			\bibitem{Zheng2024ICRA}
			H.~Zheng, Z.~J. Hu, Y.~Huang, X.~Cheng, Z.~Wang, and E.~Burdet, ``A
			user-centered shared control scheme with learning from demonstration for
			robotic surgery,'' in \emph{2024 IEEE International Conference on Robotics
				and Automation (ICRA)}.\hskip 1em plus 0.5em minus 0.4em\relax IEEE, 2024,
			pp. 15\,195--15\,201.
			
			\bibitem{Zhang2025}
			\BIBentryALTinterwordspacing
			S.~Zhang, O.~So, M.~Black, Z.~Serlin, and C.~Fan, ``Solving multi-agent safe
			optimal control with distributed epigraph form {MARL},'' in \emph{Proceedings
				of Robotics: Science and Systems}, 2025. [Online]. Available:
			\url{https://roboticsconference.org/program/papers/27/}
			\BIBentrySTDinterwordspacing
			
			\bibitem{Gu2024}
			S.~Gu, L.~Yang, Y.~Du, G.~Chen, F.~Walter, J.~Wang, and A.~Knoll, ``A review of
			safe reinforcement learning: Methods, theories, and applications,''
			\emph{IEEE Transactions on Pattern Analysis and Machine Intelligence},
			vol.~46, no.~12, pp. 11\,216--11\,235, 2024.
			
			\bibitem{Fan2024}
			K.~Fan, Z.~Chen, G.~Ferrigno, and E.~De~Momi, ``Learn from safe experience:
			Safe reinforcement learning for task automation of surgical robot,''
			\emph{IEEE Transactions on Artificial Intelligence}, vol.~5, no.~7, pp.
			3374--3383, 2024.
			
			\bibitem{Lowe2017}
			R.~Lowe, Y.~Wu, A.~Tamar, J.~Harb, P.~Abbeel, and I.~Mordatch, ``Multi-agent
			actor-critic for mixed cooperative-competitive environments,'' in
			\emph{Advances in Neural Information Processing Systems}, vol.~30.\hskip 1em
			plus 0.5em minus 0.4em\relax Curran Associates, Inc., 2017.
			
			\bibitem{Yu2022MAPPO}
			C.~Yu, A.~Velu, E.~Vinitsky, J.~Gao, Y.~Wang, A.~Bayen, and Y.~Wu, ``The
			surprising effectiveness of ppo in cooperative multi-agent games,''
			\emph{Advances in neural information processing systems}, vol.~35, pp.
			24\,611--24\,624, 2022.
			
			\bibitem{Li2022}
			Y.~Li, A.~Sena, Z.~Wang, X.~Xing, J.~Babi{\v{c}}, E.~van Asseldonk, and
			E.~Burdet, ``A review on interaction control for contact robots through
			intent detection,'' \emph{Progress in Biomedical Engineering}, vol.~4, no.~3,
			p. 032004, 2022.
			
			\bibitem{Iovene2025TBME}
			E.~Iovene, L.~Casadio, J.~Fu, F.~Costa, G.~Ferrigno, and E.~De~Momi,
			``Human-robot shared control for osteotomy procedure,'' \emph{IEEE
				Transactions on Biomedical Engineering}, vol.~72, no.~9, pp. 2698--2708,
			2025.
			
			\bibitem{Yu2021}
			T.~Yu, J.~Huang, and Q.~Chang, ``Optimizing task scheduling in human-robot
			collaboration with deep multi-agent reinforcement learning,'' \emph{Journal
				of Manufacturing Systems}, vol.~60, pp. 487--499, 2021.
			
			\bibitem{Xiao2023}
			J.~Xiao, J.~Gao, N.~Anwer, and B.~Eynard, ``Multi-agent reinforcement learning
			method for disassembly sequential task optimization based on human--robot
			collaborative disassembly in electric vehicle battery recycling,'' \emph{ASME
				Journal of Manufacturing Science and Engineering}, vol. 145, no.~12, p.
			121001, 2023.
			
			\bibitem{Shafti2020}
			A.~Shafti, J.~Tjomsland, W.~Dudley, and A.~A. Faisal, ``Real-world human-robot
			collaborative reinforcement learning,'' in \emph{2020 IEEE/RSJ International
				Conference on Intelligent Robots and Systems (IROS)}, 2020, pp.
			11\,161--11\,166.
			
			\bibitem{Ou2023}
			Y.~Ou and M.~Tavakoli, ``Towards safe and efficient reinforcement learning for
			surgical robots using real-time human supervision and demonstration,'' in
			\emph{2023 International Symposium on Medical Robotics (ISMR)}, 2023, pp.
			1--7.
			
			\bibitem{Wang2023}
			C.~Wang, C.~P{\'e}rez-D'Arpino, D.~Xu, L.~Fei-Fei, K.~Liu, and S.~Savarese,
			``Co-gail: Learning diverse strategies for human-robot collaboration,'' in
			\emph{Proceedings of the 5th Conference on Robot Learning}, ser. Proceedings
			of Machine Learning Research, vol. 164.\hskip 1em plus 0.5em minus
			0.4em\relax PMLR, 2022, pp. 1279--1290.
			
			\bibitem{Esfandiari2025TMRB}
			M.~Esfandiari, J.~W. Kim, P.~Zhang, J.~S. Heng, P.~Gehlbach, R.~H. Taylor, and
			I.~I. Iordachita, ``Bimanual robotic eye manipulation using adaptive sclera
			force control: Towards safe retinal vein cannulation,'' \emph{IEEE
				transactions on medical robotics and bionics}, 2025.
			
			\bibitem{Pacchierotti2016TBME}
			C.~Pacchierotti, D.~Prattichizzo, and K.~J. Kuchenbecker, ``Cutaneous feedback
			of fingertip deformation and vibration for palpation in robotic surgery,''
			\emph{IEEE Transactions on Biomedical Engineering}, vol.~63, no.~2, pp.
			278--287, 2016.
			
		\end{thebibliography}
	\end{document}